\RequirePackage[hyphens]{url}
\documentclass[a4paper,UKenglish,cleveref, autoref, thm-restate]{lites-v2021}

\usepackage{cleveref}
\usepackage{resources/scheduling} 
\usepackage{algorithm}
\usepackage[noend]{algpseudocode}
\usepackage{tikz}
\usetikzlibrary{shapes.geometric}
\usepackage{xcolor}

\newcommand\REV[1]{{\color{black} #1}}

\nolinenumbers

\title{A Survey of Real-Time Support, Analysis, and Advancements in ROS~2}

\author{Daniel Casini}{Scuola Superiore Sant'Anna, Pisa, Italy} {daniel.casini@santannapisa.it}{
https://orcid.org/0000-0003-4719-3631}{}%

\author{Jian-Jia Chen}{TU Dortmund University, Germany}{jian-jia.chen@tu-dortmund.de}{https://orcid.org/0000-0001-8114-9760}{}

\author{Jing Li}{New Jersey Institute of Technology, Newark, USA}{jingli@njit.edu}{
https://orcid.org/0000-0002-6865-7290}{}

\author{Federico Reghenzani}{Politecnico di Milano, Milan, Italy}{federico.reghenzani@polimi.it}{
https://orcid.org/0000-0002-1888-9579}{}

\author{Harun Teper}{TU Dortmund University, Germany}{harun.teper@tu-dortmund.de}{
https://orcid.org/0000-0003-2873-9096}{}

\authorrunning{Daniel Casini et al.} 

\Copyright{Jane Open Access and Joan R. Public} 
\funding{This work was partially supported by: the project SERICS (PE00000014) under the MUR National Recovery and Resilience Plan funded by the European Union – NextGenerationEU; the US National Science Foundation (Grant No. CNS-2340171) and Department of Energy (Award No. DE-SC0024424); the NGI Enrichers Transatlantic Fellowship Programme and National Resilience and Recovery Plan (PNRR) through the National Center for HPC, Big Data and Quantum Computing;the German Federal Ministry of
Education and Research (BMBF) in the course of the 6GEM research hub
under grant number 16KISK038; the Federal Ministry of Research,
Technology and Space (BMFTR) via the \emph{6GEM+} transfer hub under
funding references 16KIS2412. This result is part of a project (PropRT) that has received funding from
the European Research Council (ERC) under the European Union's Horizon
2020 research and innovation programme (grant agreement No. 865170).}


\category{}

\Volume{11}
\Issue{1}
\DateSubmission{}
\DateAcceptance{}
\DatePublished{}

\ArticleNo{1}

\begin{document}

\maketitle

\begin{abstract}

The Robot Operating System 2 (ROS~2) has emerged as a relevant middleware framework for robotic applications, offering modularity, distributed execution, and communication.
In the last six years, ROS~2 has drawn increasing attention from the real-time systems community and industry.
This survey presents a comprehensive overview of research efforts that analyze, enhance, and extend ROS~2 to support real-time execution.
We first provide a detailed description of the internal scheduling mechanisms of ROS~2 and its layered architecture, including the interaction with DDS-based communication and other communication middleware.
We then review key contributions from the literature, covering timing analysis for both single- and multi-threaded executors, metrics such as response time, reaction time, and data age, and different communication modes.
The survey also discusses community-driven enhancements to the ROS~2 runtime, including new executor algorithm designs, real-time GPU management, and microcontroller support via micro-ROS.
Furthermore, we summarize techniques for bounding DDS communication delays, message filters, and profiling tools that have been developed to support analysis and experimentation.
To help systematize this growing body of work, we introduce taxonomies that classify the surveyed contributions based on different criteria.
This survey aims to guide both researchers and practitioners in understanding and improving the real-time capabilities of ROS~2.
\end{abstract}

\section{Introduction}

The Robot Operating System (ROS)~\cite{ros_org} has been established as one of the most popular middleware frameworks for prototyping, developing, and deploying robotics applications. With its first release in 2007, it has become more and more widespread, thanks to a modular architecture, a considerable community support, extensive libraries, and excellent interoperability. Its impact goes beyond robotics, with applications including, among others, industrial automation, smart infrastructure, and automotive~\cite{d2023ros, kato2018autoware, nabissi2022ros,ros_industrial_cons}.
Notably, Autoware~\cite{kato2018autoware}, the largest and most popular open-source autonomous driving framework, builds upon the ROS~2 software stack to modularly integrate perception, location, planning, and control functionalities.

After about 10 years since the first release, it became clear that the original ROS was constrained by several architectural limitations and shortcomings, which were difficult to address without a complete redesign of the framework. To overcome these issues with a completely new design, ROS~2 was released in 2017.
One of the key objectives of ROS~2 was to provide support for real-time computations, thus allowing the implementation of time-critical applications~\cite{macenski2022robot}.
Therefore, the real-time capabilities of ROS~2 have drawn increasing attention from both the academic community and industry.

Since then, the real-time systems community has intensively studied the topic of guaranteeing timing constraints under ROS~2, with particular focus on ensuring the end-to-end timing constraints of \emph{processing chains} that may span from sensing to actuation through multiple software components, cores, and possibly distributed processing platforms.
Indeed, providing such guarantees is essential for safely integrating ROS~2 applications into cyber-physical systems, which are characterized by tight interaction of the processing systems with the physical world through sensing, computation, and actuation.

Since the very first work~\cite{Casini2019}, it has become evident that ROS~2 was characterized by a unique scheduling behavior that did not match any state-of-the-art scheduling algorithm previously studied in the real-time systems literature.
Therefore, over the last years, the community has largely explored the topic by proposing: 

\begin{itemize}
    \item real-time analysis methods to analyze end-to-end latency metrics such as response times, reaction times, and data age of ROS~2 applications, which are the key building blocks to enable principled design-space exploration without needing the full prototyping and deployment of the system;
    \item customized schedulers, aimed at improving the timing predictability of the default ROS~2 scheduling algorithm;
    \item tools to help the development of real-time applications using ROS~2, as well as system-level enhancements to support the interaction between ROS~2 and hardware accelerators or other frameworks; and
    \item techniques for bounding ROS-related communication delays (e.g., DDS), message filters, and profiling tools that have been developed to support analysis and experimentation.
\end{itemize}

After more than 6 years of work by the real-time systems communities towards these goals, we believe it is essential to understand what we have accomplished and how far we are from achieving the ambitious goal of supporting real-time computing in ROS~2.

This survey aims to guide both researchers and practitioners in understanding the literature about the real-time capabilities of ROS~2.
To help systematize this growing body of work, several taxonomies have been introduced in this survey to classify the surveyed contributions based on different criteria, including scheduling strategies, communication mechanisms, and timing models.

\paragraph*{Survey structure}

The discussion of this survey is divided into the following parts: \cref{sec:background} summarizes the necessary background on the real-time aspects of ROS~2, covering its internal scheduling mechanisms, layered architecture, and interaction with DDS-based communication and other middleware backends;
\cref{sec:scheduling} discusses the articles on scheduling analyses on native ROS~2. \cref{sec:enhancements} discusses novel approaches that extend the ROS~2 scheduling approaches and introduces GPU and latency management techniques; 
\cref{sec:tools} recaps the papers dealing with tools, profiling, and empirical analyses;
\cref{sec:conclusion} concludes the paper.

\section{Background}
\label{sec:background}

We start the survey by providing the necessary background. In particular, \cref{sec:ros2background} summarizes the layered architecture of ROS~2, discusses its scheduling mechanism, and reports the differences between the different executor types natively available in ROS~2.
\cref{sec:beyondros2} reports some relevant information regarding the interaction of ROS~2 with lower-level middleware that are used to dispatch messages, such as the DDS, and describes the \emph{micro-ROS} initiative, aimed at using ROS~2 on microcontrollers. Finally, \cref{sec:ros1} briefly summarizes some approaches that have been presented in the state-of-the-art to improve the real-time performance of ROS~1, so as to provide a historical note.

\subsection{ROS~2}
\label{sec:ros2background}

In this section, we provide an overview of the ROS~2 architecture, the system design, and the basic ROS~2 scheduling and communication mechanisms.

\begin{figure}[h!]
    \centering
    \includegraphics[width=0.8\textwidth]{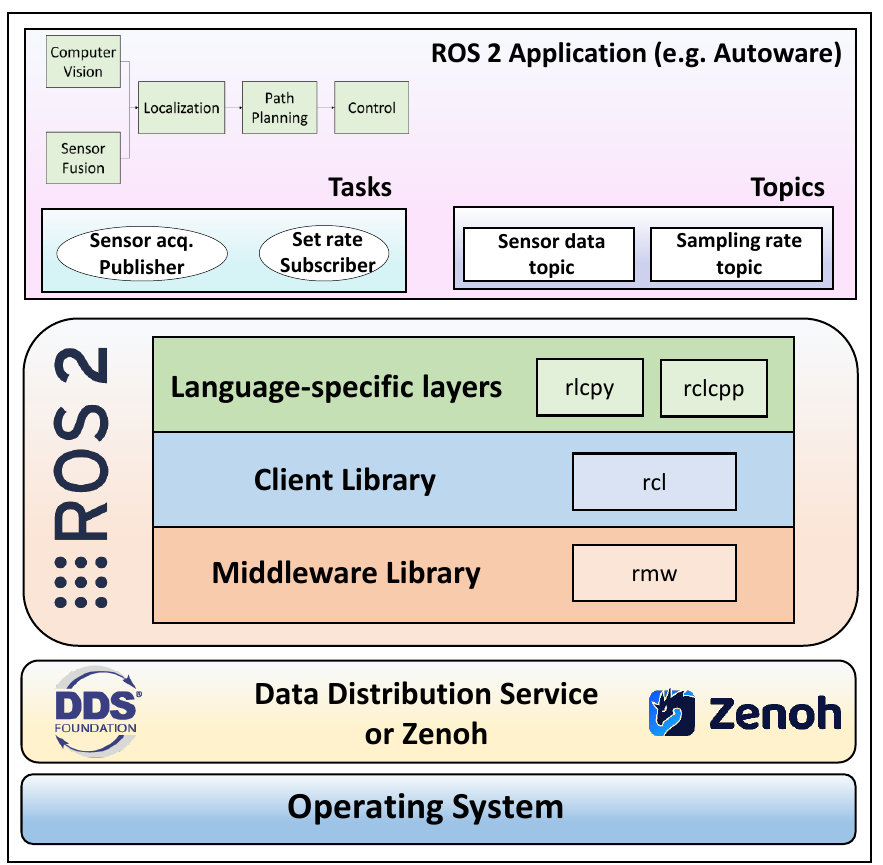} 
    \caption{ROS~2, with its layers and interactions.} 
    \label{fig:layers} 
\end{figure}

\smallskip
\noindent \textbf{Overview.} As shown in \cref{fig:layers}, ROS~2 acts as a middleware between the application and the operating system, providing a set of tools and libraries to build robotic applications, as well as interfaces for inter-process and inter-machine communication.
ROS~2 itself is composed of several layers that enable its modular architecture.
At the top, there are programming-language-specific layers: the two most popular ones are \texttt{rclpy} for Python and \texttt{rclcpp} for C++, respectively.
The literature on real-time performance of ROS~2 assumes the use of the C++ interface, since C++ is best suited for time-critical applications.
Below, the \texttt{rcl} layer provides the core functionalities of ROS~2 and the \texttt{rmw} layer provides a unified API for the underlying middleware used for message dispatching (DDS~\cite{omg_dds14} or Zenoh~\cite{Corsaro2023}).

ROS~2 applications are realized through a distributed network of components, \REV{as illustrated in~\cref{{fig:components}}}.
Each component is realized as a node, which is the fundamental unit of composition in ROS~2 systems and encapsulates a set of related functionalities.
The functionalities of a node are implemented as \textbf{callbacks}, which are functions that are executed in response to specific events.
These events are triggered by different task types, such as subscribers that receive messages (\emph{event-driven}) or timers that elapse after a specified time interval (\emph{time-driven}).

\begin{figure}[h!]
    \centering
    \includegraphics[width=0.8\textwidth]{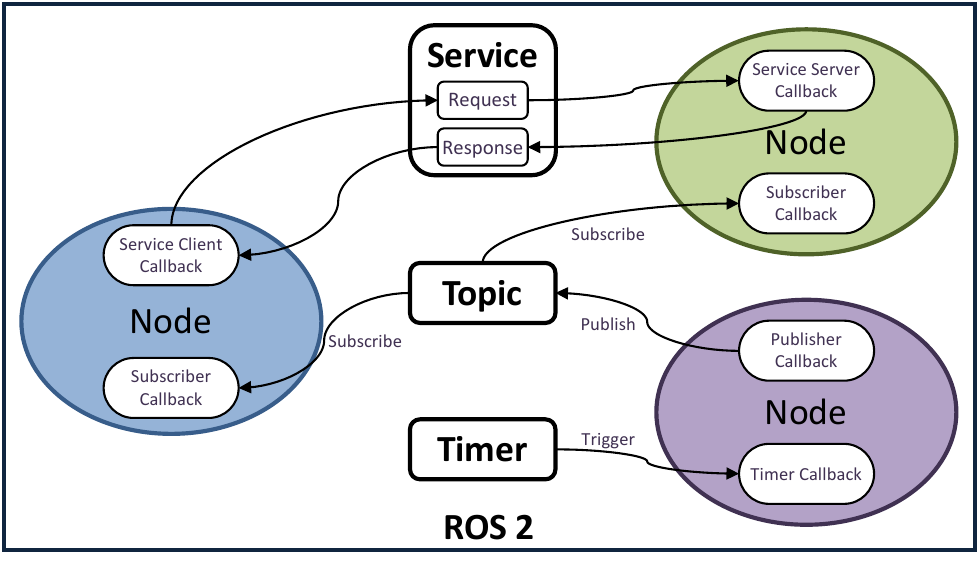} 
    \caption{\REV{ROS~2 components.}} 
    \label{fig:components} 
\end{figure}

In ROS~2, nodes can communicate with each other using the \emph{publish-subscribe} paradigm.
Each node can include publishers and subscribers that allow them to communicate via messages through so-called \emph{topics}.
When a callback \emph{publishes} on a topic, it sends a message in broadcast to all the \emph{subscribers} that subscribe to the same topic.
The reception of the message triggers a corresponding activation, which eventually leads to the execution of the subscriber callback.

Another mechanism to activate callbacks is via \emph{timers}, which trigger the execution of a callback periodically, according to a specified time interval.
Furthermore, ROS~2 also provides a mechanism for remote procedure calls (RPCs) through \emph{services}.
These feature a client-server architecture, where a client callback sends a request to a service server, which processes the request and returns a response, which is then processed by a client callback, \REV{as shown in~\cref{{fig:components}}}.

Finally, ROS~2 also includes the concept of \emph{waitables}, a general mechanism for integrating custom asynchronous events into the executor.
Waitables enable developers to define new sources of execution that can be triggered by external events, such as file descriptors or other system-level notifications, and can be seamlessly integrated in ROS~2 alongside traditional mechanisms.

\smallskip
\noindent \textbf{Scheduling in ROS~2.} In ROS~2, the callbacks are computational activities of interest for bounding real-time performance metrics, such as response times and end-to-end latencies.
However, callbacks are functions dispatched by the ROS~2 middleware and are subject to a two-level scheduling hierarchy.

On the one hand, callbacks are scheduled by ROS~2 threads, which implement their own scheduling mechanism at the level of C++ functions, as extensively discussed in the following.
On the other hand, ROS~2 threads are scheduled by the underlying Operating System (OS), which most commonly is Linux\footnote{Other OSes, such as QNX and VxWorks, FreeRTOS, are also supported.}.

The scheduling behavior of OS-level schedulers has been largely studied for several decades.
For example, both Linux and QNX provide priority-based schedulers and reservation-based schedulers, for which the timing behavior was largely explored~\cite{Dasari2022, Lelli2016}.
Conversely, ROS~2 implements custom internal scheduling mechanisms that have only been studied since 2019~\cite{Casini2019}.
As detailed below, the scheduling logic of ROS~2 is significantly different from any other scheduler that has been studied before.
In ROS~2, the scheduling algorithm is implemented within the \emph{executor}, the C++ function dispatcher at the middleware level.
Three default executor algorithms are implemented in the standard distribution: \textbf{(1)} the single-threaded executor, \textbf{(2)} the multi-threaded executor, and \textbf{(3)} the static single-threaded executor.
The static single-threaded executor is a variant of the single-threaded executor, optimized for reduced overhead, in which the dynamic task addition is not supported.

\smallskip
\noindent \textbf{Single-threaded executor in ROS~2.}
The single-threaded executor is the default callback scheduling mechanism in ROS~2.
Its behavior, outlined in \cref{alg:singlethreaded}, differs significantly from classical priority-based schedulers.
Instead of managing a global list of all ready callbacks, it manages a local set, referred to as \emph{wait set}, that is periodically updated from the underlying layers.
The executor's core logic to update the wait set revolves around a repeating cycle of two phases: polling and processing.

\begin{algorithm}
\caption{Single-threaded Executor Scheduling Pseudocode}
\begin{algorithmic}[1]\label{alg:singlethreaded}
\Function{SingleThreadedExecutorScheduler}{}
    \State Initialize \textit{wait set} as empty set
    \While{true}
        \If{\textit{wait set} is empty}
            \State \textit{wait set} $\gets$ PollCallbacks() \label{alg:singlethreaded_linepoll}
            \Comment{Polling point}
        \EndIf
        \ForAll{callback in \textit{wait set} ordered by priority}
            \State \Comment{Priority order is: timers, subscriptions, service requests, service reply, waitables} \label{alg:singlethreaded_lineselect}
            \State execute(callback) \Comment{Callbacks executed non-preemptively} \label{alg:singlethreaded_lineexec}
            \State remove(callback) from \textit{wait set}
        \EndFor
    \EndWhile
\EndFunction
\Function{PollCallbacks}{}
    \State Initialize \textit{newWaitSet} as empty set
    \ForAll{callback $c_i$ in readyCallbacks} \Comment{query readyCallbacks from the underlying layer}
        \If{\textit{newWaitSet} does not contain callback $c_i$}
            \State Add(callback) to \textit{newWaitSet}
            \Comment{Ensure unique callbacks in \textit{wait set}} \label{alg:singlethreaded_lineunique}
        \EndIf
    \EndFor
    \State \Return \textit{newWaitSet}
\EndFunction
\end{algorithmic}

\end{algorithm}

\noindent \textbf{Polling and Processing Cycle.} 
The time between two polling events is known as \emph{processing window}~\cite{Casini2019}.
During a processing window, the executor executes callbacks from its local wait set.
When the wait set becomes empty, a \emph{polling point} occurs.
At this point, the executor repopulates its wait set by querying the underlying layers for all callbacks currently ready (\cref{alg:singlethreaded}, line~\ref{alg:singlethreaded_linepoll}).
All callbacks that follow this update behavior are typically called \emph{polled callbacks}~\cite{Blass2021}.

Once the wait set is populated, the executor begins executing the stored callbacks sequentially.
The selection follows a fixed-priority, non-preemptive policy based on a two-level hierarchy:

\begin{enumerate}
    \item \textbf{Callback Type:} Callbacks are processed in a predefined order, thus implementing an implicit per-type priority ordering: timers, subscriptions, service handlers, service clients, and waitables (\cref{alg:singlethreaded} Line~\ref{alg:singlethreaded_lineselect}).
    \item \textbf{Registration Order:} Callbacks of the same type are prioritized depending on the order in which they were created and added to the executor. 
\end{enumerate}

Once selected, a callback is executed to completion non-preemptively (\cref{alg:singlethreaded} Line~\ref{alg:singlethreaded_lineexec}).
However, the executor's thread itself can still be preempted by the underlying operating system.

\begin{figure}
\centering
  \begin{tikzpicture}[yscale=0.4, xscale=0.4]

    \foreach \x in {0,...,30} {
    \draw[dashed, gray!50] (\x,0) -- (\x,5);
    }
    
    \begin{scope}[shift={(0,4)}] 
    \taskname{$\tau_1$}
    
    \timeline{0}{31}{}
    
    {\releases{0}}
    {\releases{6}}
    \draw[->, \releasearrowprops, dotted, blue] (12,0) -- (12, \releasearrowlength);

    {\releases{18}}
    {\releases{24}}
    
    {\execname{0}{3}{$C_{1,1}$}}
    {\execname{8}{10}{$C_{1,2}$}} 
    {\execname{22}{24}{$C_{1,3}$}} 

    \end{scope}

    \begin{scope}[shift={(0,2)}] 
    \taskname{$\tau_2$}
    
    \timeline{0}{31}{}
    
    {\releases{0}}
    {\releases{9}} 
    {\execname{3}{8}{$C_{2,1}$}}
    {\execname{24}{30}{$C_{2,2}$}}
    \end{scope}
    
    \begin{scope}[shift={(0,0)}] 
    \taskname{$\tau_3$}
    
    \timeline{0}{31}{}
    \labelling{0}{30}{5}{0}
    
    \releases{4} 
    \execname{10}{22}{$C_{3,1}$} 

    \draw[red, dashed, very thick] (0,0) -- (0,6);
    \draw[red, dashed, very thick] (8,0) -- (8,6);
    \draw[red, dashed, very thick] (22,0) -- (22,6);
    \draw[red, dashed, very thick] (30,0) -- (30,6);

    \draw[<->, line width=1.0pt, blue] (0,6) -- (8,6);
    \draw[<->, line width=1.0pt, blue] (8,6) -- (22,6);
    \draw[<->, line width=1.0pt, blue] (22,6) -- (30,6);

    \node[anchor=north, text=red, align=center] at (0, 8.5) {polling\\point};

    \node[anchor=south, text=blue] at (15, 6) {processing window};
    
    \end{scope}
    \end{tikzpicture}
    \caption{ROS~2 polling and processing cycle. Polling points are represented with vertical red dashed lines. The processing window length is instead represented with horizontal, double-arrowed, blue lines.}
    \label{fig:polling} 
\end{figure}
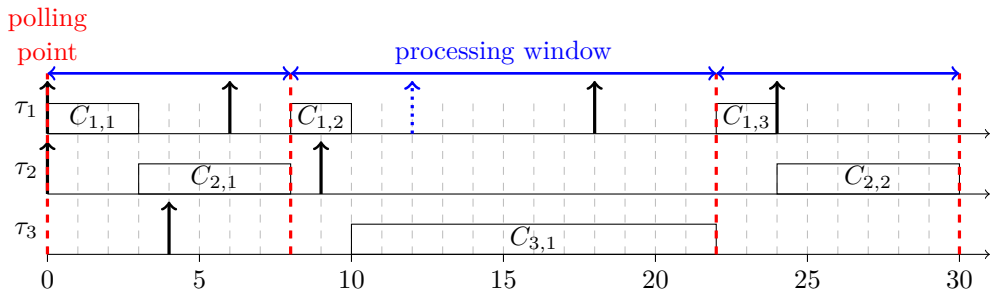

Figure~\ref{fig:polling} shows an example in which there are three callbacks: $c_1$, $c_2$, and $c_3$.
At the beginning, only $c_1$ and $c_2$ have ready callbacks (i.e., $c_{1,1}$ and $c_{2,1}$). Hence, they are sampled at the first polling point, they are added to the wait set, and they are executed.
Within the first processing window, callback instances $c_{1,2}$ of $c_1$ and $c_{3,1}$ of $c_{3}$ are released, but they are sampled only at the second polling point, thus being added to the wait set. In the second processing window, $c_{1,2}$ is executed first, followed by $c_{3,1}$. During the second processing window, two releases of $c_1$ and one release of $c_2$ occur, which are then executed in the third processing window.

Importantly, there are two critical aspects of this design that significantly affect the timing behavior of ROS~2 applications, (1) delays of lower-priority callbacks and (2) loss of releases.

\begin{itemize}
    \item If we assume lower indices to correspond to higher priorities, the polling-point mechanism causes a more serious issue in the second processing window. Indeed, instance $c_{2,2}$ is released during $c_{1,2}$'s execution; however, since $c_{2,2}$ was not ready yet at the previous polling point, it is not present in the wait set; and, when $c_{1,2}$ finishes, $c_{3,1}$ with a lower priority is selected to run instead. Since another instance of $c_1$ is released in the meantime, with higher priority, it further delays the execution of $c_{2,2}$, which is executed only later in the third processing window. 
    \item The wait set is fundamentally different than list-based ready queues used in classical real-time scheduling theory.
    In particular, the wait set enforces \emph{uniqueness} of callbacks (\cref{alg:singlethreaded} Line~\ref{alg:singlethreaded_lineunique}), i.e., it can only contain at most one instance of each ready callback in any processing window.
    This means that at each polling point, \emph{at most one instance} of each callback can be added to the wait set, regardless of how many events of a specific callback (i.e., messages for subscriptions or timer elapses) have occurred for it since the last polling point (\cref{alg:singlethreaded} Line~\ref{alg:singlethreaded_lineunique}). For instance, callback $C_1$ in Figure~\ref{fig:polling} arrives twice during the second processing window, but only one instance $C_{1,3}$ is added to the wait set and executed during the third processing window.
\end{itemize}

\smallskip
\noindent \textbf{Historical Context for Polling.}
As highlighted by Teper et al.~\cite{teper2025reconciling}, the polling mechanism of ROS~2 is significantly different from classical scheduling theory of periodic tasks, which assumes every job will be executed. Indeed, for ROS~2 versions after ``Dashing'', timer callbacks \REV{are be skipped if the executor detects that the time between the activation time of the timer callback's previous job and the start time of its next job is longer than the period of the timer callback itself}.

Compared to classical real-time scheduling, ROS~2 has the unique aspect of handling multiple types of callbacks, each with its own activation mechanism, including time-triggered and event-triggered ones.
Finally, ROS~2 also needs to provide a scheduling solution that is both capable of handling real-time critical tasks, as well as being flexible and easy to use for a wide range of robotic applications.
Thus, finding the right balance between real-time guarantees and usability has been a key design goal of ROS~2.

The exact behavior described in this section about the executor describes the behavior during the releases from 2020 (Foxy) to 2024 (Iron).
It is worth noting that this behavior has not always been consistent across all callback types.
In earlier ROS~2 distributions (up to ``Dashing''), timer callbacks were treated as \emph{privileged callbacks}~\cite{Blass2022Thesis}, i.e., they were not subject to this polling mechanism and executed normally.
Some of these cross-version changes are documented in the paper by Dust et al.~\cite{dust2023experimental}, who have conducted experiments comparing the behavior of ROS~2 Dashing to later versions. 
They show that the timer execution can be significantly delayed in newer ROS~2 distributions compared to ROS~2 Dashing due to timers being subject to the polling point and processing window approach. 
This timer issue could be alleviated by using a multi-threaded executor with dedicated executor threads for timers.
However, for complex systems, this approach may not always be feasible.

Furthermore, there have been recent developments in the distribution ROS~2 Jazzy towards completely removing the ordering inside the wait set and instead following a FIFO order~\cite{ros2_rclcpp_issue_2532}.
It remains uncertain if this behavior will be adopted and kept in future ROS~2 distributions.

\smallskip
\noindent \textbf{Multi-threaded executor.} 
The multi-threaded executor extends the polling-based principle of its single-threaded counterpart to enable parallel execution.
It utilizes a pool of threads to process callbacks concurrently.
While the fundamental process of polling for ready callbacks and populating a shared wait set remains, the multi-threaded executor can distribute these polled callbacks among its available threads, allowing them to be executed in parallel.

\smallskip
\noindent\textbf{Concurrency Management with Callback Groups.}
To manage concurrency and prevent data races when parallel execution is not desirable, ROS~2 introduces the concept of \textbf{callback groups}.
Every callback within a node is assigned to a specific callback group, which affects its execution policy.
There are two types of callback groups:
\begin{itemize}
    \item \textbf{Mutually Exclusive:} Callbacks within this type of callback groups are guaranteed not to execute in parallel with each other.
    The executor only schedules one callback from the group at a time, effectively serializing their execution, much like the single-threaded executor.
    \item \textbf{Reentrant:} This group type allows multiple callbacks assigned to it to be executed concurrently by different threads.
    It even permits multiple instances of the same callback to run in parallel if possible, e.g., if it is triggered multiple times in quick succession.
\end{itemize}

The scope of these rules is the group itself.
Callbacks belonging to different callback groups or to different nodes can always be executed in parallel, regardless of their group type, if they are assigned to the same executor.
By default, all callbacks in a node are assigned to a single, default callback group that is mutually exclusive, unless explicitly configured otherwise.
An improper assignment of callbacks to callback groups can lead to performance bottlenecks in a multi-threaded context, as callbacks that could safely run in parallel are instead forced to wait.
For performance-critical applications, developers should strategically assign callbacks to different groups or use reentrant groups to maximize parallelism and avoid unnecessary delays.

\begin{figure}[h!]
    \centering
    \includegraphics[width=0.75\textwidth]{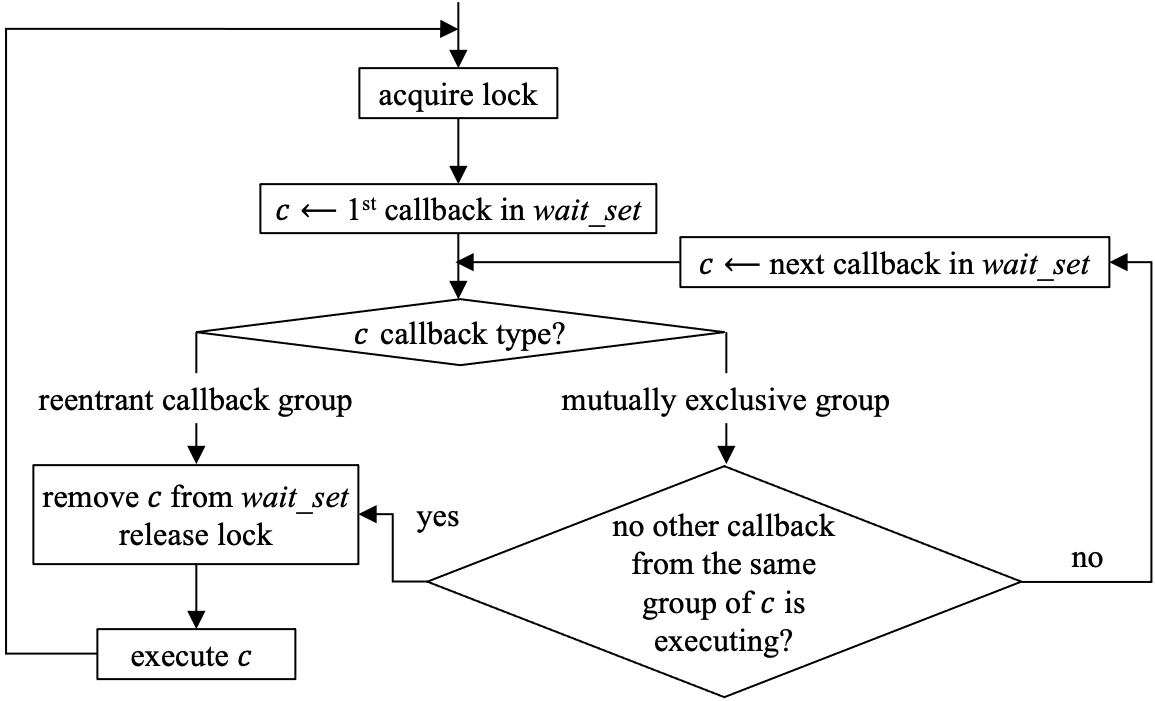} 
    \caption{Thread workflow in a multi-threaded executor} 
    \label{fig:multithread} 
\end{figure}

\smallskip
\noindent\textbf{Callback Group-Aware Polling.}
The multi-threaded executor coordinates its thread pool using locking mechanisms to ensure thread-safe access to the shared wait set. Figure~\ref{fig:multithread} summarizes the behavior of such a mechanism.
As depicted in Figure~\ref{fig:multithread}, a worker thread must first acquire a lock to gain exclusive access to the wait set.
Once it holds the lock, the thread searches for a task to execute in the wait set, following the same priority order as the single-threaded executor, but including the callback group constraints.
A ready callback in a reentrant callback group can be selected freely, while one in a mutually exclusive group can only be chosen if no other callback from that same group is currently executing.
After selecting a callback, the thread releases the lock and begins execution of the callback.
If no executable callback is found, the thread that acquired the lock waits for new callbacks to get ready, then fills the wait set again, and repeats the selection process.

\subsection{Beyond ROS~2}
\label{sec:beyondros2}

Next, we discuss the lower-level middleware used by ROS~2 to handle publish/subscribe communications and variants of ROS~2 for microcontrollers.

\smallskip
\noindent \textbf{Lower-level Pub/Sub middleware: the DDS and Zenoh.} 
The inter-process communication in ROS~2 is built upon the Data Distribution Service (DDS).
DDS is an open standard from the Object Management Group (OMG) that specifies a data-centric, publish-subscribe middleware protocol.
Its adoption in ROS~2 was driven by its proven track record in mission-critical distributed systems such as aerospace, autonomous vehicles, robotics, Internet of Things (IoT), and Industry 4.0/5.0~\cite{gambo2025systematic,  ungurean2020software, yang2012data}. Furthermore, it is also compatible with the AUTOSAR Adaptive standard~\cite{AUTOSAR_CM_2020} for automotive systems.

ROS~2 interfaces with the underlying communication system through a dedicated abstraction layer, the ROS Middleware Interface (\emph{rmw}).
This design decouples the ROS~2 client libraries from the specific middleware, allowing developers to choose the DDS implementation best suited for their application. ROS~2 supports multiple DDS implementations, each with a corresponding rmw package.
The primary options are DDS-based and involve \emph{FastDDS}, \emph{CycloneDDS}, \emph{GurumDDS}, and \emph{ConnextDDS}.
Beyond DDS, ROS~2 is also compatible with newer protocols like Zenoh~\cite{Corsaro2023}, accessible via the \emph{rmw\_zenoh package}.
Zenoh provides a more lightweight publish-subscribe and distributed query protocol.

The choice of middleware has significant implications for system performance and real-time behavior.
Because each implementation has its own internal architecture, threading model, and configuration parameters, a generic timing analysis is insufficient. For example, 
Sciangula et al.~\cite{sciangula2023bounding} provided a compositional model that generalizes beyond the specific implementation, taking into account the common aspects.
Additionally, they provide a concrete instantiation of the model for FastDDS to devise a response-time analysis. \REV{This work is further extended in~\cite{Sciangula2025} to also support all the policies of FastDDS and to include multiple communication modes, as well as integrating the analysis with ROS~2.}
Understanding and modeling the specific timing properties of the chosen middleware is crucial for accurately predicting end-to-end latencies and ensuring that real-time requirements are met.

To illustrate the complexity of the DDS configuration, we now consider FastDDS.
One of the most critical settings is the message dispatch mode, which dictates how data is sent. There are two modes:
\begin{itemize}
    \item \textbf{Synchronous}: The publishing ROS~2 callback directly handles the message-sending operation, blocking its thread until the message publication is complete.
    \item \textbf{Asynchronous}: The message-sending operation is offloaded to a dedicated service thread, allowing the publishing callback to complete without waiting for the message to be sent.
    In FastDDS, this service thread is called the flow-controller thread.
\end{itemize}

\begin{figure}
    \centering
    \begin{tikzpicture}[yscale=0.6, xscale=0.4]
        \draw[dashed, red, thick] (0,4) -- (0,6);
        \draw[dashed, red, thick] (3,4) -- (3,6);
        \draw[dashed, red, thick] (6,4) -- (6,6);
        \draw[dashed, red, thick] (9,4) -- (9,6);
        \draw[dashed, red, thick] (6,0) -- (6,2);
        \draw[dashed, red, thick] (12,0) -- (12,2);
        \draw[dashed, red, thick] (18,0) -- (18,2);
        \draw[dashed, red, thick] (24,0) -- (24,2);
        
        \begin{scope}[shift={(0,4)}] 
        \taskname{$C_1$}
        
        \timeline{0}{25}{}
        
        \execname{0}{3}{$C_{1,1}$}
        \execname{3}{6}{$C_{1,2}$}
        \execname{6}{9}{$C_{1,3}$}
        \end{scope}
        
        \begin{scope}[shift={(0,2)}] 
        \taskname{Buffer}
        
        \timeline{0}{25}{}
        
        \execname{0}{3}{\{\}}
        \execname{3}{6}{\{1\}}
        \execname{6}{9}{\{2\}}
        \execname{9}{12}{\{2,3\}}
        \execname{12}{18}{\{3\}}
        \execname{18}{24}{\{\}}
        \end{scope}
        
        \begin{scope}[shift={(0,0)}] 
        \taskname{$C_2$}
        
        \timeline{0}{25}{}
        \labelling{0}{20}{5}{0}
        
        \execname{6}{12}{$C_{2,1}$}
        \execname{12}{18}{$C_{2,2}$}
        \execname{18}{24}{$C_{2,3}$}
        \end{scope}
        
        \draw[->, thick, green!60!black] (1.5,4) -- (4.5,3);
        \draw[->, thick, blue] (4.5,4) -- (7.5,3);
        \draw[->, thick, orange] (7.5,4) -- (10.5,3);
        
        \draw[->, thick, green!60!black] (4.5,2) -- (9,1);
        \draw[->, thick, blue] (7.5,2) -- (15,1);
        \draw[->, thick, orange] (15,2) -- (21,1);

        \node[ellipse, draw, thick, align=center, text width=3cm, fill=white] at (16, 5) {Message in buffer \\ once job finishes};
    \end{tikzpicture}
    \caption{DDS-based synchronous inter-node communication. }
    \label{fig:dds_sync}
\end{figure}
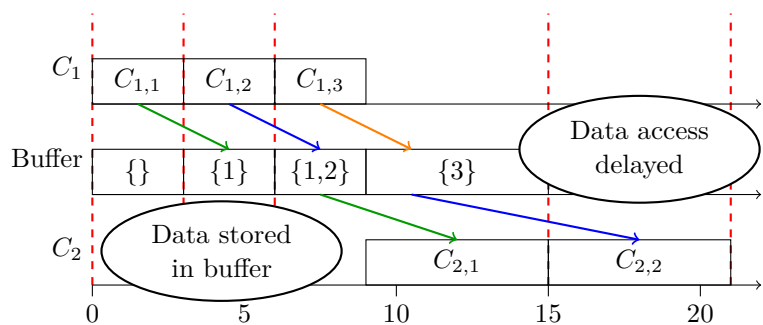

\REV{We illustrate these behaviors using the example system in Figures~\ref{fig:dds_sync} and~\ref{fig:dds_async}, which consist of a callback $C_1$ that publishes three messages to a subscriber callback $C_2$ via DDS. Each callback is run by its own executor. We assume that the executor of $C_2$ starts after time point $6$, and that each job of callback $C_1$ publishes at the end of its execution.

Figure~\ref{fig:dds_sync} illustrates the timing behavior of DDS-based synchronous communication in ROS~2.
As soon as a job of callback $C_1$ finishes its publish operation, the message is guaranteed to be stored in the buffer of $C_2$.
Since messages may arrive while the executor of $C_2$ is already processing a previous message, the buffer stores multiple messages that are then processed in FIFO order, one per processing window.

Figure~\ref{fig:dds_async} illustrates the timing behavior of DDS-based asynchronous communication for the same scenario.
In contrast to the synchronous case, the publishing callback $C_1$ does not wait for the message to be sent.
Instead, the message is handed off to the flow-controller thread, which handles the actual transmission.
As a result, the execution time of each $C_1$ instance is shorter, since it no longer includes the time spent on the publish operation.
However, the subscriber's buffer is only updated once the flow-controller thread has completed the transmission, which may introduce additional latency compared to the synchronous mode.
This is illustrated between time points $6$ and $8$, in which the buffer is empty, while message $m_2$ is still being processed by the flow-controller thread.
}

Moreover, this flow-controller thread can itself be configured with different scheduling policies, such as \texttt{HIGH\_PRIORITY} (i.e., fixed-priority), \texttt{FIFO}, or \texttt{ROUND-ROBIN}.
The choice between synchronous and asynchronous mode, combined with the scheduling policy of the flow-controller thread, can substantially impact the overall latency and determinism of message delivery in ROS~2 systems.
Furthermore, on the receiver side, messages are typically delivered by a DDS listener thread to the ROS~2 \REV{executor} in FIFO order, adding another stage to the communication pipeline that must be considered in a full system analysis.

\begin{figure}
    \centering
    \begin{tikzpicture}[yscale=0.6, xscale=0.4]
        \draw[dashed, red, thick] (0,6) -- (0,8);
        \draw[dashed, red, thick] (3,6) -- (3,8);
        \draw[dashed, red, thick] (6,6) -- (6,8);
        \draw[dashed, red, thick] (6,0) -- (6,2);
        \draw[dashed, red, thick] (12,0) -- (12,2);
        \draw[dashed, red, thick] (18,0) -- (18,2);
        \draw[dashed, red, thick] (24,0) -- (24,2);

        \begin{scope}[shift={(0,6)}] 
        \taskname{$C_1$}
        \timeline{0}{25}{}
        \execname{0}{2}{$C_{1,1}$}
        \execname{3}{5}{$C_{1,2}$}
        \execname{6}{8}{$C_{1,3}$}
        \end{scope}

        \begin{scope}[shift={(0,4)}] 
        \taskname{FC-Thread}
        \timeline{0}{25}{}
        \execname{2}{4}{$m_1$}
        \execname{4}{8}{$m_2$}
        \execname{8}{12}{$m_3$}
        \end{scope}

        \begin{scope}[shift={(0,2)}] 
        \taskname{Buffer}
        \timeline{0}{25}{}
        \execname{0}{4}{\{\}}
        \execname{4}{6}{\{1\}}
        \execname{6}{8}{\{\}}
        \execname{8}{12}{\{2\}}
        \execname{12}{18}{\{3\}}
        \execname{18}{24}{\{\}}
        \end{scope}

        \begin{scope}[shift={(0,0)}] 
        \taskname{$C_2$}
        \timeline{0}{25}{}
        \labelling{0}{24}{5}{0}
        \execname{6}{12}{$C_{2,1}$}
        \execname{12}{18}{$C_{2,2}$}
        \execname{18}{24}{$C_{2,3}$}
        \end{scope}

        \draw[->, thick, green!60!black] (1,6) -- (3,5);
        \draw[->, thick, blue] (4,6) -- (6,5);
        \draw[->, thick, orange] (7,6) -- (10,5);

        \draw[->, thick, green!60!black] (3,4) -- (5,3);
        \draw[->, thick, blue] (6,4) -- (10,3);
        \draw[->, thick, orange] (10,4) -- (15,3);

        \draw[->, thick, green!60!black] (5,2) -- (9,1);
        \draw[->, thick, blue] (10,2) -- (15,1);
        \draw[->, thick, orange] (15,2) -- (21,1);

        \node[ellipse, draw, thick, align=center, text width=3.5cm, fill=white] at (15, 7) {Send message via\\ flow-controller thread};
    \end{tikzpicture}
    \caption{DDS-based asynchronous inter-node communication.}
    \label{fig:dds_async}
\end{figure}

\smallskip
\noindent \textbf{micro-ROS.} 
micro-ROS is the official extension of ROS~2 designed to run on resource-constrained microcontrollers (MCUs).
Primarily developed by eProsima, Bosch, the FIWARE Foundation, PIAP, and Acutronic Robotics, with support from European research projects such as H-ROS~\cite{mayoralhros} and OFERA~\cite{ofera_h2020}, its goal is to bring the standard ROS~2 APIs and tools to embedded systems.

To operate within the limited memory and processing power of an MCU, micro-ROS employs a client-agent architecture.
\begin{itemize}
    \item \textbf{micro-ROS Client}: The micro-ROS client is a lightweight library that runs on the MCU.
    It implements the ROS~2 APIs but does not connect directly to the global data space.
    \item \textbf{micro-ROS Agent}: The micro-ROS agent runs on a more powerful machine (e.g., a single-board computer) and acts as a proxy.
    It connects to the main ROS~2 network via the standard DDS protocol.
\end{itemize} 

Communication between the client and agent uses the DDS-XRCE (eXtremely Resource-Constrained Environments) protocol, which is a lightweight, client-server protocol specifically designed for embedded systems with limited resources. The agent effectively bridges the resource-constrained device into the ROS~2 ecosystem.

The scheduling of callbacks in micro-ROS is designed for predictability and minimal overhead, critical for embedded applications.
The default executor provided by the \texttt{rclc} library (C-language client library), referred to as the \texttt{rclc} executor, operates as a simple, non-preemptive, static-priority scheduler.
Its execution logic is as follows:

\begin{itemize}
    \item \textbf{Static Ordering}: Like ROS~2 itself, micro-ROS also features timer-based and event-based triggering mechanisms through timers and subscriptions to initiate the executor's processing cycle. However, callbacks (e.g., timers, subscriptions) are executed in a fixed, sequential order based on the user-assigned priorities, making the system's behavior more predictable.
    \item \textbf{Subscription Semantics}: For the trigger condition of subscriptions, the developer can specify \emph{execution semantics}, configuring if the callback of a subscription should run only if new data is available (\texttt{ON\_NEW\_DATA}) or if it should run in every cycle (\texttt{ALWAYS}), which is useful for polling or fixed-rate tasks.
    \item \textbf{Triggered Execution}: On top of the basic triggering mechanisms, the executor's processing cycle can also be initiated by a trigger condition.
    This allows for the implementation of complex, event-driven control logic. Key trigger mechanisms include the following policies:
    \begin{itemize}
        \item \textbf{ANY}: Starts execution if at least one callback is ready.
        \item \textbf{ALL}: Starts execution only if all specified callbacks are ready.
        \item \textbf{ONE}: Starts when a message for a specific callback arrives.
        \item \textbf{USER-DEFINED}: Allows custom triggering logic, including hardware interrupts.
    \end{itemize}
    \item \textbf{Data Consistency Semantics}: To support time-triggered, real-time systems, the executor offers Logical Execution Time (LET)~\cite{henzinger2001embedded} semantics.
    When enabled, the executor reads and buffers all available inputs at the beginning of a cycle.
    All callbacks within that cycle then operate on this consistent, timestamped snapshot of data.
    This contrasts with the default ROS~2 semantics, where the data of a callback is fetched from the queue immediately before its execution.
\end{itemize}

For applications requiring concurrency, the \texttt{rclc} executor also provides an experimental multi-threaded model.
In this architecture, a main executor thread monitors for incoming data and handles the trigger conditions.
For each callback, a dedicated worker thread is spawned with configurable scheduling parameters provided by the underlying RTOS.
The main executor thread then dispatches new message data to the appropriate callback function, which executes in its dedicated worker thread.
This design allows developers to leverage native RTOS scheduling features such as thread priorities, scheduling policies, and advanced scheduling algorithms (including reservation-based scheduling) to manage real-time callback execution.

\subsection{Overview of the ROS 1 Approaches for Real Time}
\label{sec:ros1}

A primary motivation for the development of ROS~2 was the inherent difficulty of achieving real-time performance in its predecessor, ROS~1~\cite{kay2015, macenski2022robot}.
By design, ROS~1 nodes are implemented as standard operating system threads, meaning their execution is delegated entirely to the underlying Linux scheduler.
In typical installations, this means relying on general-purpose schedulers like the \emph{Completely Fair Scheduler (CFS)} \cite{pabla2009completely} or the more recently \emph{Earliest Eligible Virtual Deadline First (EEVDF)}~\cite{stoica1995earliest}, which are optimized for general-purpose systems.

While the Linux kernel \REV{has gained} powerful real-time capabilities, e.g., through tools for its fine-grained configuration~\cite{de2022operating, de2025timerlat}, the PREEMPT\_RT patch~\cite{de2020thread, de2017automata}, and the SCHED\_DEADLINE scheduler~\cite{phoronix2024,Lelli2016}, ROS 1 itself lacks the internal architecture and APIs needed to effectively leverage them.
In contrast, the design of ROS~2 aims at explicitly targets real-time support through a modular execution model, DDS-based communication, and configurable DDS QoS policies~\cite{perez2015modeling}.
The fundamental design limitation of ROS~1 prompted numerous research efforts to retrofit real-time capabilities onto the ROS 1 framework, which can be broadly categorized into two main strategies.

The first strategy involved architecturally isolating critical tasks using a co-kernel or multi-kernel approach to bypass the standard Linux scheduler.
For example, Delgado et al.~\cite{DELGADO20198} used the Xenomai real-time co-kernel to execute core ROS 1 components as high-priority tasks alongside the main OS.
A similar concept was explored by Wei et al.~with RGMP-ROS~\cite{wei2014rgmp-ros} and its multi-core evolution, RT-ROS~\cite{wei2016171}, which dedicated specific CPU cores to a real-time operating system to manage time-sensitive ROS nodes.

A second strategy focused on building scheduling frameworks on top of a standard Linux system.
A notable example is the ROSCH framework~\cite{Saito2018}, which introduced a DAG-based scheduling system, synchronization mechanisms, and fail-safe functions to manage ROS~1 nodes, later extending support to GPUs with ROSCH-G~\cite{Suzuki2018}.

Ultimately, these approaches were workarounds for the core issue: ROS 1 was not designed with real-time performance in mind.
The absence of proper APIs and internal structures to manage timing constraints meant that scientific contributions often focused on empirical latency measurements and mitigation, as exemplified by tools developed by Nishimura et al~\cite{nishimura2021}, rather than providing a formal scheduling analysis with provable guarantees.
This gap was a key driver for the fundamental architectural redesign that led to ROS~2.

\section{Analysis of Real-Time Behavior in ROS~2}
\label{sec:scheduling}

We start the survey by summarizing the papers on the real-time analysis of ROS~2.
Achieving predictable real-time performance in ROS~2 requires a deep understanding of the various factors that contribute to the real-time behavior.
Over the past several years, a significant body of research has emerged to formally analyze these factors from different perspectives.
This section provides a comprehensive overview of these analytical efforts.
We begin by examining the core of ROS~2's internal scheduling logic, covering works that model the behavior of both the \emph{single-threaded} (\cref{subsec:analysissinglethreaded}) and \emph{multi-threaded} executors (\cref{subsec:analysismultithreaded}).
We then explore studies that focus on the communication stack, discussing techniques for bounding delays introduced by lower-level middleware like DDS and specialized message synchronization filters in \cref{sec:comms}.
Subsequently, we explore an alternative approach based on \emph{formal verification and model checking} in \cref{subsec:formalverification}, which explores the verification of timing properties in a different way.
Finally, a \emph{taxonomy} is presented in~\cref{sec:taxonomy} to systematically classify and compare the different analytical approaches discussed throughout this section.

\subsection{Analysis of the Single-Threaded Executor}
\label{subsec:analysissinglethreaded}

\noindent \textbf{The first ROS~2 paper in real-time systems.}
The first work on the real-time analysis of ROS~2 systems was published by Casini et al.~\cite{Casini2019}.
In their paper, they described the single-threaded executor algorithm for the first time\footnote{The ROS~2 documentation now reports the behavior of the executor algorithm, based on~\cite{Casini2019}, see \\ \url{https://docs.ros.org/en/kilted/Concepts/Intermediate/About-Executors.html}}
and provided a system model suitable for analyzing ROS~2 \emph{processing chains}, which are sequences of callbacks that communicate (via topics) across multiple executors, spanning from sensing to actuation.
They modeled these chains as a directed acyclic graph (DAG) of callbacks to capture intersection and complex data flows~\cite{Casini2019}.

To analyze this model, the authors adopted \emph{Compositional Performance Analysis (CPA)}~\cite{Henia2005}, a framework for analyzing complex systems by examining individual components in isolation.
In their mapping, ROS~2 executors act as components that supply CPU time, while callbacks are computations that consume it.
This required modeling three key aspects:
\begin{itemize}
    \item \textbf{Available CPU Resources:} To abstract away the OS-level scheduler, the authors of~\cite{Casini2019} used the \emph{supply-bound function}~\cite{casini2017constant, Lipari2003, Shin2003} abstraction, denoted as $sbf_{k}(\Delta)$, which represents a lower bound on the processing time an executor $k$ receives in any given time interval of length $\Delta$, which can be derived for reservation-based schedulers like Linux's SCHED\_DEADLINE~\cite{Lelli2016}.
    \item \textbf{Workload Arrival:} Callback releases were modeled using arrival curves, denoted as $\eta_i(\Delta)$, which upper-bounds  the number of releases of a callback $c_i$ in any interval of length $\Delta$.
    Arrival curves for initial ``source'' callbacks (e.g., timers) are provided, while curves for subsequent callbacks are systematically derived by means of the arrival-curve propagation process~\cite{Henia2005}.
    \item \textbf{Workload Execution:} The execution time of each callback was modeled using the classical \emph{worst-case execution time (WCET)} abstraction~\cite{wilhelm2008worst}.
\end{itemize}
Based on this, the paper derived a formal model and  \emph{response-time analysis} for processing chains.
The analysis supports arbitrary DAG structures, including callbacks that are triggered by messages from multiple topics (an \emph{OR} activation semantic), and accounts for external event sources like device drivers.
Their baseline method computes the response-time bound by summing the individual response-time bounds of each callback in a chain.

The core of their analysis is a search for the worst-case scenario within a so-called ``busy window'', which can be informally defined as a period of continuous processing. The interested reader in the correct and formal definition is left to the original paper.
The response-time analysis presented in the paper builds on computing the response-time of arbitrary callback instances released at an arbitrary point in time $A$ within the busy window.
The per-callback response time is then computed as the maximum among all possible instances released at arbitrary values of $A$.
Clearly, this requires bounding and discretizing the space of possible time instants $A$.

First, instead of checking over an unbounded time interval (i.e., $A \ge 0$), the paper restricts the search space to the interval $[0, L)$, where $L$ is an upper bound on the length of a busy window.
Second, it discretizes this interval into a finite set of evaluation points, making the problem computationally feasible. The technique builds on previous work on the abstract response-time analysis framework~\cite{Bozhko2020}.

Finally, the authors proposed an improved holistic analysis to mitigate pessimism from the ``pay-burst-only-once'' problem~\cite{LeBoudec2001, Schliecker2009} for subchains with a single executor.
While this work laid the essential foundation for all subsequent ROS~2 analysis, its model only captured the negative scheduling effects of the single-threaded executor, paving the way for future refinements.

\smallskip

\noindent \textbf{``Second-generation'' analysis methods.}
Building on the foundational work from Casini et al.~\cite{Casini2019}, subsequent research introduced more precise analysis methods.
These ``second generation'' works focused on refining the model to capture more nuanced behaviors of the single-threaded executor, leading to tighter and more accurate latency bounds.

A key improvement by Tang et al.~\cite{Tang2020} focused on optimizing \emph{linear processing chains}, i.e., chains that cannot include branching and each callback can belong to only one chain. 
Their work improved analysis precision for these common structures and introduced a priority assignment strategy.
It demonstrated that a chain's end-to-end latency is primarily affected by the priority of its final (sink) callback, leading to the recommendation of promoting sink priorities.

In parallel, Blass et al.~\cite{Blass2021} enhanced the analysis for \emph{arbitrary DAGs} by incorporating two novel concepts:
\begin{itemize}
    \item \textbf{Execution Time Curves:} Instead of relying on a single WCET, they used execution time curves, introduced by Quinton et al.~\cite{Quinton2012}, which model the cumulative execution demand over multiple job releases.
    This better captures the variability in callback execution times common in real-world applications.
    \item \textbf{Starvation-Freedom Property:} They formally modeled the executor's round-robin-like nature, which prevents high-priority callbacks from indefinitely starving low-priority ones.
    While the original analysis treated this behavior as a source of pessimism for high-priority tasks, this work also modeled its positive impact on lower-priority callbacks, resulting in a more balanced and accurate system-wide analysis.
\end{itemize}

This approach was validated using callback graphs extracted from the real-world Turtlebot3 navigation stack.
Later, Tang et al.~\cite{tang2023real} further refined the analysis for DAGs by drawing deeper comparisons between the ROS~2 executor and classical round-robin scheduling, leveraging graph decomposition techniques to better characterize workload interference.

\smallskip

\noindent \textbf{Analysis of Reaction Time and Data Age.}
In addition to the traditional response-time metrics, recent research has introduced a shift to metrics that focus on data propagation for ROS~2.
This line of research was initiated by Teper et al.~\cite{Teper2022}, who expanded the system model to cover more diverse communication patterns.
Beyond \REV{the typical publish-subscribe mechanism} (termed \emph{inter-node} communication), their work also formalized \emph{intra-node} communication, where callbacks exchange data by storing it within the same node, and \emph{external} communication, where callbacks interact with external data interfaces.
Critically, they focus on two new performance metrics:
\begin{itemize}
    \item \textbf{Maximum Reaction Time (MRT):} The latency from an external event to the system's corresponding reaction.
    \item \textbf{Maximum Data Age (MDA):} The data freshness, in which the latency is determined backward from the actuation to the cause that originated it.
\end{itemize}
This new focus was supported by a more general cause-effect chain model that allowed chains to start with a timer, with following callbacks being either subscriptions or timers, but not allowing consecutive timers.
Despite these advances, this initial analysis was constrained to a single executor, excluding chains that span multiple executors and thus not supporting distributed systems.
Furthermore, it does not incorporate OS-level scheduling overheads, nor support for supply-bound functions to abstract the underlying OS scheduling.

The support for multi-executor systems has been introduced in a later work by the same authors~\cite{Teper2024}.
The work provided a formal upper-bound analysis for both MRT and MDA in these distributed setups, additionally lifting the prior restriction against having consecutive timers within a chain.
Beyond the analysis, the paper also introduced an optimization framework using constrained programming to minimize end-to-end latencies.
This framework automatically configures system parameters such as node-to-executor assignments, DDS communication modes (synchronous or asynchronous), task prioritization policies, and timer periods.
Evaluated on an autonomous racing software stack, the optimization reduced the analytical upper bound by up to 50.2\% and the maximum measured latency by 19.8\%.

Focusing more on fundamental theory of end-to-end latencies, Günzel et al.~\cite{gunzel2023equivalence} explored the precise relationship between the maximum reaction time (MRT) and maximum data age (MDA) metrics that Teper et al. analyzed.
The authors formally proved that, after an initial system warm-up period, MRT and MDA are equivalent.
Crucially, they demonstrated this equivalence holds under very general and non-restrictive assumptions, applying to a wide variety of systems including those with over- and undersampling, various scheduling policies (e.g., preemptive or non-preemptive, static-priority or EDF), and distributed setups.
Furthermore, they provide a case study about a ROS~2 system that shows that this condition also experimentally holds.

Orthogonally, Teper et al.~\cite{Teper2023} addressed the impact of system architecture on performance issues like message loss and high end-to-end latencies.
The authors provided two key analytical bounds.
First, they derived a bound for timer periods to prevent sensor undersampling, where a timer fails to process all available sensor data due to interference from other callbacks on the same executor.
Second, they analyzed subscription buffers, concluding that a buffer size of two is sufficient to prevent message loss for a subscription with a single synchronous publisher.
They also showed that with proper callback prioritization, this could be reduced to one.
Based on these findings, the paper proposed a heuristic for callback prioritization designed to reduce both buffer sizes and end-to-end latencies.
To enable this heuristic, the authors suggested a minor architectural change to the ROS~2 executor: prioritizing subscription callbacks over timer callbacks by swapping their sampling order.
The evaluation of these changes demonstrated a reduction in end-to-end latencies by up to 40\%.

The impact of different data handling semantics was explored by Tang et al.~\cite{tang2024timing}, who focused on \emph{data refreshing}, where limited buffer sizes cause fresh data to overwrite older messages.
They noted that while this reduces the overall workload, it can counter-intuitively worsen end-to-end latency in some scenarios compared to the standard First-In-First-Out (FIFO) approach, necessitating a dedicated analysis.
The paper provides a formal timing analysis to calculate upper bounds for both maximum reaction time and data age, specifically for chains using data refreshing, considering how message overwriting affects data propagation.
Their method offers more precise bounds than existing FIFO-based analyses, which they show are safe but pessimistic.
Furthermore, the authors prove that end-to-end latency can be optimized by setting the input buffer size of the first regular callback in the chain to one.

\smallskip

\noindent \textbf{Jitter Control and the LET Paradigm.} While most analyses focus on bounding the worst-case end-to-end latency, a predictable system also requires low jitter.
Control systems rely on low jitter in order to provide reliable and consistent timing.
To address this, Abaza et al.~\cite{abaza2024managing} proposed a non-intrusive method called \textit{latency shaping}, inspired by the \emph{Logical Execution Time (LET)}~\cite{henzinger2001embedded} paradigm, that uses table-driven reservation servers to achieve near-zero jitter.
The core idea of their approach is to decouple when a task finishes from when its output is released, ensuring that data is always published at a predictable, fixed point in time.
The authors use a two-server design to implement this method in ROS~2 without modifying application code:

\begin{itemize}
    \item \textbf{High-Priority Server:} This server runs the main chain computations, ensuring that the processing is completed as quickly as possible.
    \item \textbf{Low-Priority Server:} This server has a precisely timed slot dedicated solely to publishing the chain's final output at a deterministic point in time.
\end{itemize}
This separation guarantees that even if the computation time varies, the final output is always sent with near-zero jitter.
For chains using synchronous publishing, the authors architecturally extend the chain with a \textit{republisher} node to maintain this separation without modifying application code.
This technique also supports creating multiple \textit{latency bands}, which allows a system to operate in different modes (e.g., city vs. highway driving) with a predictable latency for each mode.

\smallskip
\REV{In addition to the aforementioned methods, new research directions are emerging. For example, Lee et al.~\cite{lee2022probabilistically} and Han and Kim~\cite{han2023minimizing} propose probabilistic real-time analysis and optimization methods for end-to-end latency in autonomous-driving task graphs, using Autoware as a case study; however, these works target more general scheduling assumptions and are not specific to the ROS~2 scheduling policy.}

\subsection{Analysis of the Multi-Threaded Executor}
\label{subsec:analysismultithreaded}

Analyzing the real-time behavior of the multi-threaded executor is inherently more complex due to the increased concurrency and resource contention.
Research in this area has focused on developing accurate response-time analyses that address these challenges.

The initial response-time analysis for the multi-threaded executor was presented by Jiang et al.~\cite{jiang2022real}.
Their work provided two different scheduling tests characterized by different accuracy versus complexity trade-off for ROS~2 processing chains under multi-threaded executors. The second one leverages a dynamic programming approach to explore the space of interfering sequences, thus providing better schedulability results in the evaluation. Conversely, the first one is simpler but more pessimistic.
The experiments also identified that a careful assignment of callback groups for callbacks is essential to avoid an increased response time when switching from single to multi-threaded executors.
The authors also observed that concurrency issues arise when callbacks share a common resource, potentially leading to higher response times than in the single-threaded executor case. In particular, when several callbacks belong to the same mutually exclusive group, a low-priority callback can be repeatedly removed from and re-added to the wait set together with new high-priority instances. Every time this occurs, the new high-priority callbacks are chosen first, so the low-priority callback is postponed again, increasing the response time. With a single-threaded executor, this does not occur because once a callback is in the queue, it is not removed and reinserted.  
Teper et al.~\cite{teper2024thread} showed that this concurrency issue makes the multi-threaded executor prone to starvation.
Specifically, they observed that, since high-priority callbacks in a callback group can be re-added to the wait set, higher-priority jobs would then be executed before the lower-priority jobs of the callback group that were previously added, thus causing starvation.
They provided a few configurations that end up in starving tasks in ROS~2 systems, uncovering unobserved cases in the previous two works on response-time analysis for the multi-threaded executor~\cite{jiang2022real, sobhani2023timing}.
To address the problem, the authors proposed a design change to the multi-threaded executor in ROS~2, which prevents the refresh of the wait set for higher-priority tasks in a callback group.
They formally and experimentally proved the absence of starvation scenarios, guaranteeing that both the single-threaded and multi-threaded executors align with respect to starvation-freedom.
Sobhani et al.~\cite{sobhani2023timing} also proposed a response time analysis for the multi-threaded executor, but compared to \cite{jiang2022real}, their model accepts arbitrary deadlines and considers fixed-priority scheduling, which makes the multi-threaded executor follow the priority of the corresponding chain when scheduling each callback.

\subsection{Bounding Communication and Synchronization Delays}
\label{sec:comms}

The real-time performance of applications developed using ROS~2 is often not only affected by the scheduling mechanism of ROS~2 itself. As discussed in Section~\ref{sec:beyondros2}, lower-level middleware, such as the DDS, may also have a substantial impact on the timing performance since they are in charge of delivering messages according to publish-subscribe. This section first discusses them, and then presents several works related to how ROS~2 computational activities can selectively process incoming messages.

\smallskip
\noindent\textbf{\REV{Intra-Process Communication and} Lower-Level Middleware (DDS).}
In ROS~2, communication occurs in two main ways: \emph{intra-process} and \emph{inter-process} communication.
Intra-process communication, which handles message exchanges within the same executor, is optimized to avoid unnecessary copies when sending messages from publishers to subscribers.

\REV{Intra-process communication has been recently characterized by Luo et al.~\cite{luo2025scalability}, who showed that its overhead is not constant and can vary based on the message configuration, workload structure, and usage semantics. To improve scalability, the paper proposes guidelines for configuring intra-process communication and an allocation-free message pooling mechanism.} 

In contrast, inter-process communication manages the more complex task of connecting publishers and subscribers across different executors, often on separate machines, and relying on the underlying DDS middleware.
Because the performance of distributed robotic systems hinges on the efficiency of this layer, most real-time research has focused on analyzing the overheads of inter-process communication.

A key factor in this analysis is the message exchange mode, which can be either \emph{synchronous} or \emph{asynchronous}~\cite{sciangula2023bounding}.
In synchronous mode, a publishing callback blocks its executor thread until the message is received by all subscribers, ensuring immediate delivery but potentially introducing significant delays.
Meanwhile, in asynchronous mode, the callback delegates the sending process to the DDS middleware and continues its execution without blocking.
The complexities of this asynchronous mode were formally analyzed by Luo et al.~\cite{luo2023modeling}, who provided an upper-bound for communication delay when the DDS WriterHistory of ROS~2 is configured with either the FIFO (First-In-First-Out) (used in  ROS~2 Humble Hawksbill) or InterestTree policy (used in ROS~2 Foxy Fitzory).
The FIFO policy ensures that messages are delivered to subscribers in the order they were sent, while the InterestTree policy maintains separate FIFO queues for each publisher, sending the messages of each publisher sequentially, and ensuring fair delivery.

For a more comprehensive model of the DDS and an end-to-end analysis for distributed applications, Sciangula et al.~\cite{sciangula2023bounding} developed a detailed, compositional DDS model using the CPA framework~\cite{Henia2005}.
Their work offers a generic DDS model and a concrete instantiation for FastDDS, one of the most common implementations. Their model captures the behavior of key DDS components in asynchronous mode, including the \emph{flow-controller threads} that dispatch messages and the \emph{listener threads} that 
receive them, allowing these delays to be incorporated into a full system analysis.
This understanding of the FastDDS implementation was also leveraged by Teper et al.~\cite{Teper2024} to provide an upper-bound analysis for both intra-process and inter-process communication.

Building on these detailed models, subsequent research has focused on system-level optimization and resolving cross-layer priority issues.
Sciangula et al.~\cite{sciangula2024end} developed an optimization algorithm that uses analysis-based heuristics to configure the entire communication pipeline, including thread-to-machine and thread-to-core assignments, the number of flow control threads, message partitioning, and thread priorities.
Further work~\cite{stevanato2023virtualized} presented a virtualized architecture for the DDS, leveraging Xen and Linux, focusing on two design alternatives: raw sockets bypassing the network stack and standard UDP.

\smallskip
\noindent\textbf{Message Filters and Synchronizers.}
Beyond the core communication stack, ROS~2 provides \emph{message filters}, which are libraries that allow nodes to selectively process incoming messages based on specific criteria, such as timestamps or content.
For instance, the \emph{Time Sequencer} filter enforces chronological order by discarding any message with a timestamp earlier than the last received one.

The most critical use of these filters is in \emph{Message Synchronizers}, which are essential for sensor fusion tasks where a callback must only be triggered after receiving a coherent set of inputs from multiple topics.
ROS~2 offers three main synchronization policies: \emph{ExactTime}, \emph{ApproximateTime}, and \emph{LatestTime}.
As the name suggest, the \emph{ExactTime} policy is a hard filter which verifies that the messages have the exact timestamp.

The \emph{ApproximateTime} policy is the most widely analyzed.
It groups messages that arrive within a specified time window, tolerating a small time disparity.
Research on this policy has evolved over several years.
Li et al.~\cite{li2022worst} first modeled and analyzed this time disparity, initially for systems with unlimited buffers and later extending the analysis to more realistic limited-size buffers~\cite{li2023modeling}.
Moreover, the \emph{ApproximateTime} policy potentially needs messages to remain in the message synchronizer in order to work, introducing additional latencies.
Li et al.~\cite{li2023modeling} were the first to formally model and experimentally measure these latencies, introducing the concepts of \emph{passing latency} and \emph{reaction latency}.
The \emph{passing latency} is defined as the time a message spends in the synchronizer before being output, while the \emph{reaction latency} is the time from message arrival to output, including any necessary discarding of older messages.
In 2023, Li et al.~\cite{li2023worst} provided a formal model and experimental measurements of these latencies.
Later, Wu et al.~\cite{wu2024improving} built upon~\cite{li2023worst} to provide more accurate estimates.

\REV{A notable limitation of the \emph{ApproximateTime} policy is that it requires prior knowledge of incoming message timing characteristics, and any inaccuracies in this knowledge can significantly degrade performance~\cite{sun2023seam}.
Two distinct approaches circumvent this limitation by not relying on a priori information about the messages, though they differ fundamentally in their design goals and guarantees.

The first is SEAM~\cite{sun2023seam}, a custom message synchronizer that selects the earliest-arriving message from each input queue such that the time disparity among the selected messages remains within a user-specified threshold, thereby guaranteeing well-bounded time disparity without requiring any prediction of future message arrivals.

The second is the \emph{LatestTime} policy, a more recent addition to ROS~2, which instead prioritizes maximizing output frequency.
It achieves this by implementing a \textit{zero-order-hold} mechanism, storing the latest message from each topic and publishing a complete set of the latest message per topic whenever any single topic receives a new message.
However, unlike SEAM, \emph{LatestTime} does not guarantee bounded time disparity.
Wu et al.~\cite{wu2024modeling} were the first to model the latency of this policy and uncovered a critical flaw in its design that could lead to unbounded latencies under certain scenarios.}

Finally, the impact on real-time metrics of different sensor fusion patterns has been recently explored by Sobhani et al.~\cite{sobhani2025fusionpatterns} in the context of more abstractly modeled autonomous driving systems.
\REV{Here, the authors identified task types and fusion patterns, dependent on the number of inputs and the activation patterns, and evaluated their impact on the real-time performance of the system.}

\subsection{Formal Verification with Model Checking}
\label{subsec:formalverification}

A different line of work has focused on formal verification rather than analytical timing models.
Dust et al.~\cite{dust2025pattern} proposed an approach for verifying ROS~2 applications using the UPPAAL model checker.
Their key contribution is a pattern-based method that uses reusable Timed Automata (TA) templates to simplify the formal modeling of ROS~2 components, including callbacks, communication channels, and two different versions of the single-threaded executor (ExV1 from ROS~2 Dashing and ExV2 from ROS~2 Eloquent and later releases, which differs in that in ExV1 the polling of timer callbacks is done continuously, and in ExV2 it is performed only at polling points).
The verification focuses on properties such as callback latency and buffer overflow.
Their framework supports both a holistic approach, modeling entire processing chains, and an individual-node approach, which abstracts communication to analyze nodes in isolation.
A key advantage of this method is its ability to exhaustively explore the system's state space and generate counterexample traces, which can reveal potential design flaws or critical execution paths not easily discovered through traditional testing.

This formal verification approach was later extended to the multi-threaded executor by the same authors~\cite{dust2024uppaal}.
They provide UPPAAL Timed Automata (TA) templates to model the behavior of the multi-threaded executor, including its locking mechanisms and both mutually exclusive and reentrant execution modes.
A key novelty of their work is the ability to model the influence of the underlying operating system by incorporating templates for reservation-based scheduling.
This allows for the verification of callback blocking and latency not just from the middleware's internal scheduling, but also from OS-level preemption and resource constraints.

\subsection{Taxonomy of Analysis Methods}
\label{sec:taxonomy}

To provide a general overview of the large body of work discussed in this section, we provide three summary tables, each targeting a different aspect of real-time analysis in ROS~2.
The tables are placed within their respective subsections to provide immediate context, but are described here together to give a complete overview.

\begin{itemize}

    \item \textbf{\cref{tab:ros2-analysis-comparison} (Analysis and Verification Methods)} provides a broad comparison of the analytical and formal verification papers discussed in \cref{subsec:analysissinglethreaded,subsec:analysismultithreaded,subsec:formalverification}.
    It classifies each work based on key criteria such as the executor type, system model, timing metrics, and the assumptions made about execution times, arrivals, and the underlying OS.
    The criteria are as follows:
    \begin{itemize}
        \item \textbf{Executor type} (EX.): \emph{single-threaded} (ST) vs. \emph{multi-threaded} (MT).
        \item \textbf{System model} (Model): arbitrary \emph{ Directed Acyclic Graphs} (DAG), \emph{linear chains} (Linear), or formal models based on \emph{Timed Automata} (TA).
        \item \textbf{Timing metric} (Metric): \emph{end-to-end response time} (E2E-RT), \emph{maximum data age/reaction time} (DA/RT), end-to-end timing \emph{Jitter}, or worst-case response time of callbacks (\emph{WCRT}, called individual callback latency in~\cite{dust2025pattern}).
        \item \textbf{Execution time model} (ET): \emph{worst-case execution time} (WCET), \emph{execution time curves} (ET curv.), or measured values from profiling (Measured).
        \item \textbf{External arrival model} (Arrivals): leveraging \emph{timer-based periodicity} (Timers), based on \emph{arrival curves} (Arr.curve), or a mix of \emph{periodic and sporadic} (Per./Spor.).
        \item \textbf{DDS communication}: DDS managed as a single \emph{parameter} (Para), as two distinct parameters for \emph{synchronous/asynchronous} cases (2 Para.), or no accounting of DDS (None).
        \item \textbf{Operating system model} (OS): using a \emph{supply-bound function} (SBF), using \emph{reservation-based servers} (Server), or \emph{not considered} (None).
    \end{itemize}
    \item \textbf{\cref{tab:dds_communication} (DDS Communication Analysis)} focuses specifically on the papers that analyze the DDS middleware, as detailed in \cref{sec:comms}.
    This table highlights the communication mode, the specific DDS implementation targeted, and the primary focus of the analysis, from delay modeling to system-wide optimization.
    \item \textbf{\cref{tab:message_sync} (Message Synchronization Analysis)} summarizes the research on message filters and synchronizers, discussed in \cref{sec:comms}.
    It categorizes papers by the synchronizer policy they analyze and key contributions, e.g., modeling time disparity or identifying design flaws.
\end{itemize}

\begin{table*}
\centering
\caption{Comparison of real-time analysis and verification methods for ROS~2 executors.}
\begin{tabular}{|l|l|l|l|l|l|l|l|}
\hline
\textbf{Paper} & \textbf{EX.} & \textbf{Model} & \textbf{Metric} & \textbf{ET} & \textbf{Arrivals} & \textbf{DDS} & \textbf{OS} \\
\hline
\multicolumn{8}{|c|}{\textbf{Analytical Methods}} \\
\hline
Casini et al.~\cite{Casini2019} (2019) & ST & DAG & E2E-RT & WCET & Arr.curve & Para. & SBF \\
Tang et al.~\cite{Tang2020} (2020) & ST & Linear & E2E-RT & WCET & Arr.curve & None & SBF \\
Blass et al.~\cite{Blass2021} (2021) & ST & DAG & E2E-RT & ET curv. & Arr.curve & Para. & SBF \\
Tang et al.~\cite{tang2023real} (2023) & ST & DAG & E2E-RT & WCET & Arr.curve & None & SBF \\
Teper et al.~\cite{Teper2022} (2022) & ST & Linear & DA/RT & WCET & Timers & None & None \\
Teper et al.~\cite{Teper2024} (2024) & ST & Linear & DA/RT & WCET & Timers & 2 Para. & None \\
Tang et al.~\cite{tang2024timing} (2024) & ST & Linear & DA/RT & WCET & Arr.curve & None & SBF \\
Abaza et al.~\cite{abaza2024managing} (2024) & ST & Linear & Jitter & Measured & Timers & 2 Para. & Server \\
Jiang et al.~\cite{jiang2022real} (2022) & MT & Linear & E2E-RT & WCET & Timers & None & None \\
Sobhani et al.~\cite{sobhani2023timing} (2023) & MT & Linear & E2E-RT & WCET & Arr.curve & None & SBF \\
\hline
\multicolumn{8}{|c|}{\textbf{Formal Verification Methods (Model Checking)}} \\
\hline
Dust et al.~\cite{dust2025pattern} (2025) & ST & TA & WCRT & WCET & Per./Spor. & 2. Para. & None \\
Dust et al.~\cite{dust2024uppaal} (2024) & MT & TA & WCRT & WCET & Per./Spor. & None & Server \\
\hline
\end{tabular}
\label{tab:ros2-analysis-comparison}
\end{table*}

\begin{table}[H]
\centering
\caption{Communication Analysis in ROS~2, either intra-process or DDS-based.}
\label{tab:dds_communication}
\renewcommand{\arraystretch}{1.2}
\begin{tabular}{|l|c|c|p{5.25cm}|}
\hline
\textbf{Paper} & \textbf{Communication} & \textbf{DDS Type} & \textbf{Analysis Focus} \\
\hline
Luo et al.~\cite{luo2025scalability} & Intra-process & N/A & Scalability analysis \\
\hline
Luo et al.~\cite{luo2023modeling} & Async. inter-process & Generic DDS & Worst-case communication delay modeling for FIFO \& InterestTree policies \\
\hline
Sciangula et al.~\cite{sciangula2023bounding} & Async. inter-process & FastDDS & End-to-end latency analysis with FastDDS threads modeling \\
\hline
Sciangula et al.~\cite{Sciangula2025} & Sync+Async. inter-process & FastDDS & End-to-end latency analysis with FastDDS threads modeling, and integration with ROS 2 analysis \\
\hline
Teper et al.~\cite{Teper2024} & Intra \& inter-process & FastDDS & Formal upper-bound analysis for both communication types using FastDDS \\
\hline
Sciangula et al.~\cite{sciangula2024end} & Inter-process & FastDDS & Thread-to-core assignment optimization with analysis-based heuristics \\
\hline
\end{tabular}
\end{table}

\begin{table}[H]
\centering
\caption{Message Synchronization and Filters Analysis in ROS~2}
\label{tab:message_sync}
\renewcommand{\arraystretch}{1.2}
\begin{tabular}{|l|c|c|p{5.25cm}|}
\hline
\textbf{Paper} & \textbf{Synchronizer} & \textbf{Analysis Focus} & \textbf{Key Contributions} \\
\hline
Li et al.~\cite{li2022worst} & ApproximateTime & Time disparity model & Worst-case time disparity analysis for unlimited buffer size \\
\hline
Li et al.~\cite{li2023modeling} & ApproximateTime & Buffer size constraints & Extended time disparity analysis for limited buffer sizes \\
\hline
Li et al.~\cite{li2023worst} & ApproximateTime & Latency analysis & Formal modeling of passing latency and reaction latency \\
\hline
Wu et al.~\cite{wu2024improving} & ApproximateTime & Improved analysis & Reduced pessimism in ApproximateTime latency bounds \\
\hline
Sun et al.~\cite{sun2023seam} & SEAM (Custom) & Novel synchronizer & New synchronizer without a priori knowledge, guaranteeing well-bounded time disparity\\
\hline
Wu et al.~\cite{wu2024modeling} & LatestTime & Zero-order hold & LatestTime policy analysis with unbounded latency defect identification\\
\hline
\end{tabular}
\end{table}

\section{Enhancements and Modifications for Real-Time Performance}
\label{sec:enhancements}

While the previous section focused on analyzing the real-time behavior of default ROS~2 systems, a parallel area of research has been dedicated to actively modifying and extending ROS~2 to improve its real-time performance.
This section surveys these enhancements, which range from fundamental changes to the core scheduling logic to the addition of new system-level management capabilities, as well as heterogeneous integration with domain-specific platforms.

In \cref{sec:custom_executors}, we review the design of \emph{customized executors}, a research direction that replaces the default schedulers with alternatives that are either completely new or based on classical real-time scheduling algorithms, such as fixed-priority (FP) or earliest deadline first (EDF)~\cite{liu1973scheduling}.

Next, in \cref{sec:system_level_enhancements}, we explore \emph{system-level enhancements} that address challenges beyond pure CPU callback scheduling.
This includes frameworks for managing shared resources like \emph{GPUs and hardware accelerators} to prevent common issues like priority inversion and provide end-to-end timing guarantees for heterogeneous computations.
We also cover systems for \emph{automatic latency management} that dynamically adjust ROS~2 configurations at run-time to meet high-level performance goals.

Finally, in \cref{sec:heterogeneous_integration}, we examine frameworks for heterogeneous integration, which bridge ROS~2 with specialized platforms like the automotive-grade AUTOSAR AP and the determinism-focused Lingua Franca, enabling a smoother transition from research to production environments.

\subsection{Customized Executor Design}
\label{sec:custom_executors}


The first work in this area was \emph{PiCAS} by Choi et al.~\cite{choi2021picas}, who proposed a single-threaded executor based on a fixed-priority, chain-aware policy designed to minimize end-to-end latency.
This work includes a formal latency analysis and a corresponding priority assignment scheme, based on the timing requirements and criticalities of the system. The paper showed how the standard executor algorithm can jeopardize the schedulability of time-critical callbacks, and proposes to adopt a fixed-priority scheduling algorithm within a single-threaded executor to improve real-time performance. Additionally, the paper proposes a chain-aware node allocation algorithm to minimize interference between chains. This seminal paper on customizing the executor's behavior to favor real-time predictability sparked a long series of new executor proposals.

Building directly on this foundation, Sobhani et al.~\cite{sobhani2023timing} extended the priority-driven approach to the multi-threaded executor, providing an analysis that accounts for the performance effects of mutually-exclusive callback groups.
It first analyzed the behavior of the default multi-threaded executor scheduling algorithm, then it extended it to priority-driven scheduling, allowing to analyze chains with both arbitrary and constrained deadlines.

In parallel with these fixed-priority designs, several works have explored dynamic-priority scheduling.
Arafat et al.~\cite{arafat2022response} first introduced Earliest Deadline First (EDF) scheduling by modifying the single-threaded executor, replacing its ready set with a priority queue.
Furthermore, they provide a response-time analysis for the modified executor.
In subsequent work by Arafat et al.~\cite{arafat2024dynamic}, they extended this concept to the multi-threaded executor by implementing the global EDF scheduling policy~\cite{liu1973scheduling} to enable parallel execution.
Furthermore, a response-time analysis for ROS~2 callbacks under the presented executor is proposed, which also accounts for delays due to shared resources. The paper also establishes model equivalence between the proposed executor (when considered without callback groups) and the fixed-preemption point EDF~\cite{Zhou2019}, enabling reuse of the corresponding response-time analysis.

As introduced in \cref{sec:background}, the polling-based behavior of the default ROS~2 executor is very different from traditional fixed-priority or dynamic-priority schedulers for periodically released tasks.
Rather than replacing the executor entirely, Teper et al.~\cite{teper2025reconciling} sought to bridge the gap between ROS~2's unique processing-window behavior and classical real-time theory by making only minor modifications to the recently introduced ROS~2 events executor.
In ROS~2, the default events executor introduces an events queue and allows scheduling decisions to be made immediately after a job completes.
In order to make this design more aligned with classical real-time scheduling, the authors proposed to implement a priority queue for scheduling released jobs and to manage job release and scheduling through separate queues.
This way, their design is analytically equivalent to a classical non-preemptive, work-conserving, priority-based scheduler.
Furthermore, their approach supports both fixed-priority and dynamic priority policies and yields tighter, more predictable analytical bounds and real-world performance compared to the default ROS~2 executor.

To achieve fully-preemptive scheduling, a goal not addressed by the previous non-preemptive designs, Wilson et al.~\cite{wilson2025physics} developed a modified ROS~2 executor that assigns each callback to its own dedicated thread that is then managed by Linux's SCHED\_DEADLINE scheduling class.
This callback-per-thread design supports preemptive EDF scheduling and is particularly suited for mixed-criticality systems.
Specifically, when a callback is first invoked, the executor creates a thread, sets its scheduling parameters (period, runtime, and deadline) using the SCHED\_DEADLINE scheduling class, and associates it with the callback.
To reuse threads for subsequent executions of the same callback, the executor maintains a mapping between callbacks and their threads and passes a reference to the callback's triggering event (e.g., timer or subscriber) to the associated thread.
However, their implementation requires that callbacks are non-reentrant to ensure that only one instance of a callback can execute at any given time.
Furthermore, it also assumes that each topic is published by only a single publisher per period.

Focusing on the limitations of the default multi-threaded executor, Liu et al.~\cite{liu2024rtex} identified that the lock-protected wait\_set is a major bottleneck that causes blocking and context switches.
Additionally, due to the wait\_set only being updated when it becomes empty, high-priority callbacks may experience arbitrary delays, leading to interference in end-to-end latency for time-sensitive chains.
To address this issue, they proposed \emph{RTeX}, a new multi-threaded executor design that replaces the wait\_set with a lock-free, atomic-based ready list.
This design allows multiple threads to access ready callbacks concurrently, significantly improving efficiency and timing predictability.
Furthermore, RTeX supports priority scheduling across chains, ensuring that the callback in the highest-priority chain is always selected.
It also integrates a timer monitor that checks for expired timers after each callback execution, updating the ready list in real time based on priority.
This design reduces blocking overhead, eliminates unnecessary context switches, and ensures prompt and priority-respecting scheduling, resulting in improved real-time performance for ROS~2 applications.

Addressing the prioritization for the multi-threaded executor, Wu et al.~\cite{wu2024deadline} proposed a finer-grained chain-instance-level scheduling EDF scheduling scheme for the multi-threaded executor.
Unlike traditional chain-level scheduling, their approach allows finer-grained priority assignment at the level of individual chain instances.
Specifically, chain instances are prioritized based on their deadlines, with earlier deadlines receiving higher priority.
Each callback instance inherits the priority of its corresponding chain instance, and executor threads always schedule the highest-priority callback available.
The authors also provide a response time analysis model for the proposed chain-instance-level EDF scheduling strategy.

Building on this to address mixed-criticality workloads, they later proposed a \emph{Hybrid Scheduling Executor (HSE)}~\cite{wu2024hybrid} to address the challenge of managing mixed workloads with both critical (real-time) and ordinary (best-effort) tasks, maximizing the performance of ordinary tasks while guaranteeing that all real-time constraints for critical tasks are met.
The HSE integrates two distinct schedulers: a real-time scheduler that uses the EDF strategy for critical chains, and a fair scheduler that uses a round-robin policy for ordinary chains.
To prevent interference, the HSE introduces two types of threads: exclusive threads, which are dedicated solely to executing critical tasks, and shared threads, which can run tasks from both schedulers to improve CPU utilization.
To complement this executor, the authors also developed the Chain-Aware Thread Mapping Strategy (CATMS), an approach for mapping the HSE's threads to a given number of CPU cores.

\REV{In 2025}, Ishikawa et al.~\cite{ishikawa2025work} proposed the middleware-transparent ROS~2 executor, which establishes a persistent one-to-one mapping between callbacks and Linux threads, superseding the executor's behavior and allowing callbacks to be executed directly by the OS's scheduling policy.
\REV{Very recently, Liu et al.~\cite{liu2025rosrt} proposed ROS$^{\text{RT}}$, a new execution paradigm for ROS~2 leveraging multiple worker threads that supports preemptive scheduling and supports both fixed-priority and EDF scheduling for arbitrary workloads.}

\smallskip
\REV{\noindent \textbf{micro-ROS.}
Finally, these principles of priority-driven scheduling have also been adapted to resource-constrained environments such as \emph{micro-ROS}.
The default executor in micro-ROS, known as the RCLC Executor, suffers from two critical issues highlighted in~\cite{wang2023tide}: due to its FIFO-based data processing and batch-based callback scheduling, it may violate priority ordering, leading to priority inversion during callback execution and data processing.
Additionally, the executor enters a blocking state during data transmission, leading to inefficient CPU utilization and wasted resources.
To address these issues at the system level, Wang et al.~\cite{wang2024improving} proposed PoDS, a \emph{Priority-Driven chain-aware Scheduling}  system that integrates the Timing-Deterministic and Efficient (TIDE) executor introduced in~\cite{wang2023tide}.
The TIDE Executor introduces two priority-aware queues: a timer queue sorted by timestamps and a ready queue sorted by priority.
It also incorporates a priority-based data-processing mechanism and a parallel transmission model to minimize blocking and reduce priority inversion.
These modifications make callback execution and data handling more deterministic and efficient.
Experimental results in~\cite{wang2024improving} confirm that PoDS significantly outperforms the default micro-ROS in terms of runtime predictability and real-time stability, making it a promising advancement for low-resource robotic systems.}

\smallskip
\noindent \textbf{Taxonomy of customized executors, compatibility, and current status.} \cref{tab:custom_executors} summarizes the main customized executors discussed in this section.
The table reports, for each contribution, whether the executor is single-threaded (ST) or multi-threaded (MT), the scheduling algorithm implemented, and the main context or key features.

\cref{tab:custom_executors_status}, instead, reports the current implementation status of the customized executors.
For each work, the table reports the date of the last public code commit, the targeted ROS~2 distribution, and the corresponding repository link.

\begin{table}[h!]
\centering
\caption{Overview of Customized Executors for ROS~2}
\label{tab:custom_executors}
\renewcommand{\arraystretch}{1.2}
\begin{tabular}{|l|c|c|p{5.25cm}|}
\hline
\textbf{Paper} & \textbf{Executor} & \textbf{Scheduling Algorithm} & \textbf{Context/Key Features} \\
\hline
Choi et al.~\cite{choi2021picas} & ST & Fixed Priority (FP) & Priority-driven chain-aware scheduling with latency analysis \\
\hline
Sobhani et al.~\cite{sobhani2023timing} & MT & Fixed Priority (FP) & Priority-driven chain-aware scheduling for multi-threaded executors \\
\hline
Teper et al.~\cite{teper2025reconciling} & ST & FP \& EDF & Modified events executor for non-preemptive periodic task scheduling \\
\hline
Liu et al.~\cite{liu2025rosrt} & MT & FP \& EDF & Callbacks dispatched to OS worker threads, preemptive (no wait set)\\
\hline
Wilson et al.~\cite{wilson2025physics} & MT & Preemptive EDF & Physics-informed preemptive mixed-criticality scheduling using Linux \\
\hline
Liu et al.~\cite{liu2024rtex} & MT & Priority-driven & Lock-free atomic executor for chain-aware priority-based scheduling \\
\hline
Wu et al.~\cite{wu2024deadline} & MT & Chain-instance EDF & Chain-instance level scheduling with deadline-based priorities \\
\hline
Wang et al.~\cite{wang2024improving} & ST & Priority-driven & Priority-driven chain-aware scheduling for micro-ROS\\
\hline
Arafat et al.~\cite{arafat2022response} & ST & EDF & Dynamic priority scheduling with a priority queue implementation \\
\hline
Arafat et al.~\cite{arafat2024dynamic} & MT & Global EDF & Dynamic priority scheduling for multi-threaded executors \\
\hline
\end{tabular}
\end{table}

\begin{table}[h!]
\centering
\caption{Status of Customized Executors in ROS~2. The last commit refers to the time of writing (October 2025).}
\label{tab:custom_executors_status}
\renewcommand{\arraystretch}{1.2}
\begin{tabular}{|p{2.5cm}|c|c|p{5.7cm}|}
\hline
\textbf{Paper} & \textbf{Last commit} & \textbf{ROS~2  Version}  & \textbf{Repository link} \\
\hline
Choi et al.~\cite{choi2021picas}\newline Sobhani et al.~\cite{sobhani2023timing} & February, 2023 & \emph{Eloquent} &  \url{https://github.com/rtenlab/ros2-picas} \\
\hline
Teper et al.~\cite{teper2025reconciling} & March, 2025 & \emph{Rolling} & \url{https://github.com/tu-dortmund-ls12-rt/ros2_executor_evaluations} \\
\hline
Wilson et al.~\cite{wilson2025physics} & February, 2025 & \emph{Foxy} & \url{https://github.com/RTIS-Lab/ROS-Phys-MC}\\
\hline
Liu et al.~\cite{liu2024rtex} & June, 2024 & \emph{Humble}  & \url{https://github.com/ESLab2012/RTeX}\\
\hline
Wu et al.~\cite{wu2024deadline} & December, 2024 & Unknown & \url{https://github.com/Wuzhengda55/CIL-EDF-for-ROS2-multi-threaded-Executors} \\
\hline
Wang et al.~\cite{wang2024improving} &  October, 2025 & Micro-ROS & \url{https://github.com/ESLab2012/PoDS/tree/main} \\
\hline

Arafat et al.~\cite{arafat2022response} & July, 2025 & \emph{Foxy} & \url{https://github.com/RTIS-UCF/ros_edf/} \\
\hline

Arafat et al.~\cite{arafat2024dynamic} & March, 2025 & Unknown & \url{https://github.com/RTIS-Lab/ROS-Dynamic-Executor} \\ \hline
Liu et al.~\cite{liu2025rosrt} & May, 2025 & \emph{Humble} & \url{https://github.com/sizheliu-unc/rclcpp} \\
\hline

\end{tabular}
\end{table}

\subsection{System-Level Enhancements}
\label{sec:system_level_enhancements}

Next, we discuss papers addressing inter-process communication in ROS~2, its interaction with GPU and accelerators, its integration in containerized systems and orchestration mechanisms, and mechanisms for automated latency management.

\smallskip
\noindent\textbf{Optimizing Inter-Process Communication.}
Beyond CPU scheduling, researchers have proposed system-level enhancements targeting the performance and predictability of ROS~2's inter-process communication (IPC), often focusing on limitations within the default DDS middleware or exploring alternatives.
Kim et al.~\cite{kim2025cros} identified the lack of priority propagation across ROS~2's multi-layered architecture, leading to priority inversion where low-priority DDS or kernel threads delay high-priority application messages.
To address this, they developed the \emph{Cros-Rt} framework, which explicitly aligns thread priorities across the application, middleware, and kernel layers to ensure consistent communication across layers that uphold priority ordering.
Orthogonally, Ishikawa et al.~\cite{ishikawaaso2025ros2agnocastsupporting} proposed the \emph{ROS~2 Agnocast} framework to optimize the data transfer efficiency within ROS~2 by using zero-copy shared memory for IPC.
They observed that current applications are often bottlenecked by data serialization and copying overheads in DDS, especially for large messages, but also identified key limitations in existing zero-copy DDS implementations, such as the inability to handle unsized message types (e.g., images or point clouds).
They introduce a novel publish/subscribe solution through Agnocast using shared memory and pointer-swapping that supports unsized message types, enabling true zero-copy IPC and further enhancing communication efficiency through latency reduction and throughput improvement.
Specific middleware protocols for large data objects have been investigated also by Peeck et al.~\cite{Peeck2021, Peeck2023} in the context of the DDS.
\REV{Luo et al.~\cite{luo2025flexible} proposed a multi-stage approach for enabling efficient zero-copy inter-process communication in ROS 2, with a focus on supporting dynamically structured data fields. The approach builds upon the mini memory management system component, allowing multiple processes to access the same message instance from a shared memory space, and the message propagation is adapted, which ensures compatibility with the existing ROS 2 communication framework.}

\smallskip
\noindent \textbf{Enhancing GPU and Accelerator Management.}
ROS~2 serves as a powerful resource manager for CPU tasks, but it lacks built-in support for GPU resource management. \REV{Furthermore, the vast majority of the discussed works focus on CPU scheduling only and neglect the presence of hardware accelerators. Nevertheless, }
many critical functionalities in autonomous systems, such as perception and intelligent control, require intensive GPU computation.
However, under default ROS~2, these components must submit their GPU workloads independently, without coordination.

As shown by Ali et al.~\cite{ali2025necessity} for a ROS~2 based drone system, this unmanaged approach can lead to severe performance degradation and unpredictable behavior.
They identified critical issues arising from naive utilization of CPU cores and GPUs: tasks experience undue scheduling delays when accessing busy compute hardware resources, interference occurs when concurrently running workloads contend for shared resources and slow each other down, and tasks miss deadlines due to blocked and interrupted GPU accesses.
This lack of real-time scheduling guarantees makes it difficult to provide end-to-end timing guarantees for chains that use accelerators.

To address these challenges, several frameworks have been proposed to manage accelerator access at the middleware level.
One such solution, ROSGM~\cite{li2023rosgm}, was proposed as an extension to ROS~2, introducing customizable GPU management policies and enabling dynamic switching between different policies at runtime.
ROSGM allows developers to implement various GPU scheduling strategies as plug-ins and consists of three key components: (1) a registration table for GPU callbacks, (2) a waiting queue pool that organizes requests based on the selected policy, and (3) a high-priority scheduler thread that processes the requests.
A dynamic policy switching mechanism is implemented using ROS~2's publish-subscribe model, and the framework supports three empirically evaluated policies: \emph{exclusive}, \emph{sharing}, and \emph{preemption}.

An alternative architecture for managed accelerator access was proposed by Enright et al.~\cite{enright2024paam} with the Priority-driven Accelerator Access Management (PAAM) framework, which treats accelerators as a schedulable service.
Instead of having callbacks invoke hardware directly, PAAM uses a standalone ROS~2 executor that acts as a dedicated accelerator resource server.
Client callbacks send requests to this server, which arbitrates access based on the priorities of the originating chains, thus resolving the priority inversion problem inherent in direct, unmanaged access.
PAAM employs a two-level scheduling hierarchy: requests are first mapped to hardware-level priority ``buckets'' (e.g., CUDA streams) and then scheduled within each bucket according to their fine-grained chain priority.
To minimize communication overhead, the framework uses a separate data plane with zero-copy shared memory for large kernel data, while small control messages are sent via DDS.
The work also provides a worst-case response time analysis for chains with accelerator segments and demonstrated up to a 91\% reduction in end-to-end latency for critical chains.

Beyond arbitrating access to accelerator computation, another line of work has focused on optimizing the efficiency of data movement between CPUs and accelerators.
While ROS~2 provides zero-copy semantics for multi-core CPUs, Hazcat~\cite{bell2023hardware} extends this capability to heterogeneous computing platforms.
It provides a heterogeneity-aware memory management framework that enables portable zero-copy semantics across devices by eliminating unnecessary data movement.
Hazcat introduces specialized allocators that automatically perform copy operations only when required in a decentralized manner.
Its implementation uses a shared message queue with reference counting to manage memory that may have multiple copies for different devices, with garbage collection triggered when all subscribers have processed the message.

\smallskip
\noindent \textbf{ROS with Containers, in the Edge, Fog, and Cloud.} The first efforts on leveraging ROS with containers have been spent for ROS 1: in this context, Aldegheri et al~\cite{Aldegheri2020} presented a toolchain based on container-based packaging in which applications are orchestrated across heterogeneous hardware architectures. The paper uses Docker and Kubernetes and designs the framework to be used in the context of ROS, also discussing communication mechanisms among distributed and containerized nodes. A more recent effort is due to Ichnowski et al.~\cite{Ichnowski2023}, who presented FogROS2. FogROS2 extends the ROS~2 launch system, allowing the deployment of ROS~2 nodes in the cloud and in the fog, automating the configuration (including DDS), provisioning, and data compression so that computationally intensive components can be transparently offloaded to nearby or remote servers while preserving the standard ROS~2 programming model. It uses Docker containers, and it is compatible with AWS EC2 and Kubernetes.

 Betz et al.~\cite{Betz2024} proposed a containerized microservice architecture for ROS~2–based autonomous driving software by redesigning Autoware into a set of Docker microservices orchestrated with lightweight Kubernetes (k3s). The paper compares different settings: bare-metal, single-container, and multi-container deployments on both x86 and Arm platforms, evaluating end-to-end latency, jitter, and resource usage using DDS benchmarks, a perception pipeline, and full closed-loop Autoware simulation. Their results show that containerized deployments can match or even improve end-to-end latency and jitter compared to bare-metal, while reducing CPU/memory usage and startup time, supporting the viability of microservice architectures for real-time robotic stacks.

Very recently, Betz et al.~\cite{Betz2025} presented an optimization framework for containerized ROS~2 autonomous driving software, targeting the reduction of end-to-end latency across multiple cause–effect chains in an autonomous racing stack. The paper formulates a constraint-optimization problem that jointly decides the assignment of nodes to executors, processes, and containers, whether to use DDS or intra-process communication, and possibly suitable timer periods by leveraging the timing analysis in~\cite{Teper2024}. The approach is evaluated on TUM Autonomous Motorsport’s ROS~2 Humble–based racing software, deployed as Docker microservices orchestrated via docker-compose and tested in a closed-loop Unity simulation with LiDAR/RADAR sensors.

\smallskip
\noindent \textbf{Automatic Latency Management.}
Applying formal real-time theory often requires detailed, static knowledge of the system, such as worst-case execution times and message arrival patterns.
This approach is sometimes impractical for ROS~2 applications, which rely on reusable ``black box'' components whose activation semantics, functional interactions, message arrival patterns, and worst-case execution times are either not well understood or subject to dynamic changes.
To address this challenge, ROS-Llama~\cite{blass2021automatic} was developed as an automatic latency manager that provisions ROS~2 applications dynamically at runtime.
Instead of requiring complex, low-level parameters, ROS-Llama uses a declarative approach where the user only specifies high-level latency goals for each critical processing chain and a degradation policy that manages priorities in case of transient overloads.
It operates through introspection, automatically extracting a system model at runtime, estimating the necessary parameters, and adjusting the scheduling configurations (using Linux's SCHED\_DEADLINE) as conditions evolve to decide whether any chains should be degraded to best-effort mode.
An evaluation on a TurtleBot3 demonstrated that ROS-Llama provides superior latency control under load compared to both the default Linux CFS scheduler and a criticality-monotonic SCHED\_RR baseline. Very recently, Enright et al.~\cite{enright2025theoryguided} proposed a Latency Management Executor (LaME), which leverages SCHED\_DEADLINE applied to an extended multi-threaded executor, and determines thread budget of individual executor threads, chain-to-threadclasses (a threadclass is a group of executor threads) allocation, and considers an affinity-based scheduling in which chains can execute either on a single threads, according to partitioned scheduling, or to multiple threads, according to constrained global scheduling.

A different approach to achieving predictable execution focuses on the choice of programming language and its concurrency model.
The increasing popularity of the Rust programming language for robotics has driven interest in its real-time capabilities, as its asynchronous programming paradigm differs significantly from the thread-based model of C++ ROS~2.
Skoudlil et al.~\cite{Skoudlil2025lookros2applications} explored the execution model of R2R, a community-supported asynchronous Rust client library for ROS~2.
They compared various asynchronous Rust runtimes (like Tokio and futures) and proposed a specific application structure designed for deterministic, real-time operation, which involves separating the event-sampling and callback-execution tasks into different threads with distinct priorities.
Their experiments showed that this proposed structure significantly outperforms other configurations and can achieve bounded response times that align with theoretical real-time analysis.

\subsection{Heterogeneous Integration}
\label{sec:heterogeneous_integration}

This section discusses how ROS~2 has been integrated with other frameworks and languages, such as AUTOSAR Adaptive and Lingua Franca.

\smallskip
\noindent\textbf{AUTOSAR Integration.}
A key challenge in deploying autonomous systems is bridging the significant ``platform gap'' between academic research and industrial automotive production.
While researchers favor the open-source and flexible ROS~2 for rapid prototyping, the automotive industry relies on standardized platforms like AUTOSAR Adaptive Platform (AP)~\cite{bellassai2025ap, 
furst2016autosar} to meet the strict requirements for safety, security, and real-time performance necessary for commercial vehicles.
To bridge this gap, Iwakami et al.~\cite{iwakami2024autosar} proposed a collaboration framework that enables communication between ROS~2 and AUTOSAR AP applications, particularly in cloud-based development environments for Software-Defined Vehicles (SDV).
The framework addresses the fundamental communication incompatibility between AUTOSAR AP's SOMEIP protocol and ROS~2's DDS middleware by implementing a bridge converter that operates as a ROS~2 node, enabling bidirectional data exchange through protocol conversion and service discovery mechanisms.
As a result, it enables seamless use of ROS~2 development tools for visualization and debugging, like RViz and ROSbag, with AUTOSAR AP applications, facilitating rapid prototyping in cloud environments while maintaining compatibility with automotive industry standards.
The bridge converter incorporates SOMEIP-SD functionality for service registration and discovery, performs type conversion between the different data formats, and leverages CommonAPI and vsomeip for industry-standard SOMEIP implementation.
A key innovation is the Bridge Generator, a GUI-based tool that automatically generates conversion code and configuration files from minimal user input with support for complex message types.
Evaluation on AWS EC2 instances demonstrated communication latency scaling from 5 ms for small messages to approximately 80 ms for 4 MB payloads, with throughput exceeding 40 MB/s, which is sufficient for high-resolution LiDAR sensors like the VLS-128.

\smallskip
\noindent\textbf{Lingua Franca Integration.}
Another approach addresses the inherent nondeterminism of ROS~2's standard publish-subscribe model by integrating it with Lingua Franca (LF)~\cite{menard2023high}, a deterministic coordination language designed for real-time and cyber-physical systems.
Kwok et al.~\cite{kwok2025hprm} introduced the High-Performance Robotic Middleware (HPRM), a framework built on top of Lingua Franca, which leverages LF's reactor model to provide deterministic event processing through logical time coordination.
It offers both centralized and decentralized coordination strategies where the Runtime Infrastructure (RTI) manages global event ordering or federates coordinate directly through peer-to-peer communication with Safe-to-Process (STP) offsets.
The middleware incorporates three key optimizations: an in-memory object store (Plasma) for zero-copy transfer of payloads larger than 64KB, adaptive serialization that uses out-of-band techniques for large data structures like NumPy arrays while maintaining in-band serialization for smaller data types, and an eager protocol with real-time sockets that pre-allocates 64KB buffers and disables Nagle's algorithm to minimize handshake latency.
As a result, it overcomes the high communication latency in ROS~2 when handling large data payloads.
Benchmark results demonstrate HPRM achieves up to 173$\times$ lower latency than ROS~2 when broadcasting large messages to multiple nodes, with particularly significant improvements for 10MB camera images (77$\times$ faster) and 50MB payloads (173$\times$ faster).
In real-world validation using CARLA autonomous driving simulations with parallel reinforcement learning agents and YOLO object detection, HPRM achieved 91.1\% lower latency compared to ROS~2, demonstrating its effectiveness in complex, multi-modal robotic applications requiring deterministic real-time performance.

\section{Tools, Profiling, and Experimentation}
\label{sec:tools}

While the aforementioned theoretical methods are necessary to provide formal real-time guarantees for ROS~2 systems, real-world systems also require robust evaluation of their performance through comprehensive tooling for tracing, latency measurement, and real-time benchmarking.
These tools enable visibility into message flows, callback scheduling, and end-to-end delays under diverse workloads and deployment scenarios.
In Section~\ref{sec:profilingtools}, we review profiling and tracing frameworks that instrument ROS~2 middleware and applications.
We then examine application-level latency studies leveraging these tools to characterize communication and computation overheads across hardware platforms and runtime configurations in Section~\ref{sec:applications}.

\begin{table}[h!]
\centering
\caption{Overview of Profiling Tools for ROS~2}
\label{tab:tools}
\renewcommand{\arraystretch}{1.2}
\begin{tabular}{|p{2.6cm}|p{3.5cm}|p{2.5cm}|p{4.3cm}|}
\hline
\textbf{Paper} & \textbf{Key Features} & \textbf{Compatibility} & \textbf{Source Code} \\
\hline
Bédard et al.~\cite{bedard2022ros2_tracing} (ros2\_tracing) & Instrument and analyze using ROS~2 tracepoints and LTTng & Rolling, Dashing, Eloquent, Foxy, Galactic, Humble & \url{https://gitlab.com/ros-tracing/ros2_tracing}  \\
\hline
Bédard et al.~\cite{bedard2023message} & Message flow analysis using ros2\_tracing and Eclipse Trace Compass & Rolling, Dashing, Eloquent, Foxy, Galactic, Humble & \url{https://github.com/christophebedard/ros2-message-flow-analysis} \\
\hline
Li et al.~\cite{li2022autoware_perf} (Autoware Perf) & End-to-end latency analysis using ros2\_tracing & Foxy, Galactic & \url{https://github.com/azu-lab/ROS2-E2E-Evaluation} \\
\hline
Abaza et al.~\cite{abaza2024trace} &  Tracing and timing analysis using eBPF & Foxy & Not available \\
\hline
Kuboichi et al.~\cite{kuboichi2022caret} (CARET) & End-to-end latency measurement using LTTng with additional tracepoints & Galactic, Humble, Iron, Jazzy & \url{https://github.com/tier4/caret} \\
\hline
He et al.~\cite{he2023tilde} (TILDE) &  Dynamic message
tracking using TILDE embedded nodes & Humble & \url{https://github.com/tier4/TILDE} \\
\hline
\end{tabular}
\end{table}

\subsection{Profiling Tools}
\label{sec:profilingtools}

Bédard et al.~\cite{bedard2022ros2_tracing} developed \texttt{ros2\_tracing}, a low-overhead tracing tool that enables multipurpose instrumentation for ROS~2, capturing key events such as message publications, callbacks, services, executor states, and lifecycle states.
It achieves this through tracepoint calls for both initialization and runtime events, allowing for detailed execution analysis of applications.
The tool employs the \texttt{low-overhead Linux Trace Toolkit: Next Generation} (LTTng) tracer as the default tracing backend and leverages the \texttt{tracetools} package to support tracepoint calls.
To evaluate the overhead introduced by \texttt{ros2\_tracing}, experiments measured the time between publishing a message and its reception by the subscription callback with and without the tool.
Results indicate that the mean latency overhead remains below 15\%, with each tracepoint call introducing overhead at the microsecond level.

The \texttt{ros2\_tracing} tool was later extended to extract and visualize message flows across distributed ROS~2 systems~\cite{bedard2023message}.
This extension builds an intermediate execution database from trace data collected in distributed systems, detailing ROS~2 objects (nodes, publishers, subscriptions, timers) and events (message publication, reception instances, and callback executions).
The extended tool can automatically detect one-to-one and one-to-many causal links between received and published messages for direct links and uses user-level annotations to identify more complex, indirect causal relationships.
To illustrate how the extended tool can analyze ROS~2 and application-level logic, two experiments were performed.
The first experiment used the Autoware reference system, splitting its nodes across multiple executables on two hosts within the same network.
The second experiment used the RTAB-Map simultaneous localization and mapping (SLAM) system and distributed it across two computers connected over a wireless network.

Li et al.~\cite{li2022autoware_perf} developed \texttt{Autoware Perf}, a tracing and performance analysis framework for ROS~2 applications, including Autoware.Auto.
It extends \texttt{ros2\_tracing} by allowing users to selectively trace specific nodes by manually inputting information about chosen nodes and their callback dependencies.
It aims to analyze end-to-end latency by measuring node execution time and publish/subscribe communication time.
Since \texttt{Autoware Perf} does not support tracing individual messages, the end-to-end latency, accounting for lost messages during measurement, is analytically estimated using the convolution integral of the discretized probability distribution.

Abaza et al.~\cite{abaza2024trace} proposed a tracing and measurement framework for ROS~2-based applications.
The framework utilizes an \texttt{extended Berkeley Packet Filter} (eBPF)-based tracer, which probes various functions within the ROS~2 middleware and extracts their arguments or return values to analyze the data flow within applications.
By integrating event traces from both ROS~2 and the operating system, the framework constructs a directed acyclic graph that visualizes ROS~2 callbacks, their precedence relationships, and associated timing attributes.
To evaluate its effectiveness, experiments were conducted using a real-world benchmark implementing LIDAR-based localization in Autoware's Autonomous Valet Parking system.

In addition to measuring end-to-end latency, \texttt{CARET}~\cite{kuboichi2022caret} was developed to facilitate detailed tracking and analysis of message flow in ROS~2 applications.
By adding additional tracepoints related to the message flow, \texttt{CARET} can detect lost messages that never reach their destination, thereby enabling more accurate calculation of end-to-end latency.
This tool has been used in experiments with Autoware.Universe to identify system bottlenecks and understand their underlying causes.

To support online tracking of message flow and deadline monitoring during system execution, \texttt{TILDE}~\cite{he2023tilde} was developed as a dynamic message tracking infrastructure for ROS~2 applications.
\texttt{TILDE} augments the ROS~2 system by publishing specialized messages called MessageTrackingTags alongside the original ROS~2 messages to log the publishing and subscribing relationships between nodes.
A special deadline detector node is also added at the end of the data flow to collect these tags, computes end-to-end latencies, and reports deadline misses.

\subsection{Profiling Various Latencies using Different Applications}
\label{sec:applications}

The first work exploring the performance of ROS~2 was by Maruyama et al.~\cite{Maruyama2016}, who described the differences between ROS 1 and ROS~2 and evaluated ROS~2 performance using an alpha version available in 2016.
Kronauer et al.~\cite{kronauer2021latency} conducted empirical experiments on desktop PCs and Raspberry Pis to profile communication latency in ROS~2 across multiple hardware nodes.
Their experiments focused on a use case where a sensor reading is published in one node and then processed through a chain of nodes, with variables including publishing frequency, payload size, number of computing nodes, and quality of service (QoS) settings.
They measured the 95th percentile of end-to-end latency in ROS~2 distributed systems using default settings and different DDS middleware, including Connext, FastRTPS, and CycloneDDS, focusing specifically on latency attributed to ROS~2 rather than the DDS middleware itself.
To capture intra-layer communication latency, they employed the \texttt{ros2profiling} package to obtain precise timestamps.
The results reveal that end-to-end latency decreases as frequency increases, especially with longer data-processing pipelines.
This effect may be due to unknown energy-saving features in other hardware components or changes in thread scheduling priorities.
For payload sizes exceeding the UDP fragmentation threshold, latency increases with payload size.
The most significant sources of latency were attributed to DDS and rclcpp notification delay.
Notably, ROS~2 introduces up to 50\% additional latency over raw DDS for small messages (128 B), while latency for larger messages (500 KB) is largely dominated by the DDS middleware.
Furthermore, DDS middleware performance differs between Raspberry Pis and desktop PCs, with Raspberry Pis exhibiting higher latency and greater fluctuation in latency compared to desktop PCs.

Betz et al.~\cite{betz2023fast} conducted latency measurements using the \texttt{ros2\_tracing} tool and hardware-in-the-loop (HiL) simulation of \texttt{Autoware.Universe}, an open-source autonomous driving software stack.
Their study focused on the impact of timer frequencies, executor priorities in ROS~2, DDS implementations, operating system (OS) scheduling, and various hardware configurations.
The evaluated path in Autoware starts from receiving LiDAR data and ends at the controller output via EKF localization, involving 12 nodes and 15 callbacks, including 3 timer callbacks.
Performance metrics measured include end-to-end latency of the evaluated path, communication latency between ROS~2 nodes, compute latency for callback processing, and idle latency due to intra-node timers.
Results indicate that prioritizing subscription callbacks over timer callbacks increases both idle time and end-to-end latency.
The findings suggest that increasing timer frequencies, as computational resources allow, can reduce idle latency.
While different DDS implementations show minimal impact on communication latency, variations such as whether separate DDS processes are created do influence callback execution, leading to differences in end-to-end latency.
Additionally, different OS schedulers achieve relatively similar throughput, though the choice of round-robin quantum affects end-to-end latency.
Lastly, experiments show that hyperthreading and variable CPU clock speeds increase end-to-end latency due to delays in executing timer callbacks.

Barut et al.~\cite{barut2021benchmarking} conducted a communication benchmark involving node/component communication typical in robotics to compare the real-time performance (in terms of message latency and jitter) of ROS~2 and the earlier Open Robot Control Software (OROCOS) framework.
The benchmark involved periodically sending a request with echo data that was immediately responded to upon receipt.
Experiments were performed on both vanilla and PREEMPT\_RT patched Linux kernels under normal conditions and stress scenarios induced by the standard Linux \texttt{stress} tool.
The results show that, under vanilla Linux without stress, ROS~2 and OROCOS exhibit comparable message latencies, although ROS~2 shows higher jitter.
Under stress, OROCOS experiences significant latency spikes.
In comparison, under vanilla Linux with stress, ROS~2 suffers a linear increase in latency of sending messages periodically, likely due to dropped messages under high load.
When using the PREEMPT\_RT patched kernel, ROS~2 performance improves markedly, showing bounded latencies.
These findings highlight the critical importance of stress testing in evaluating system performance, particularly for real-time applications in robotics.

\section{Conclusions}
\label{sec:conclusion}

In this survey, we reviewed the state of the art on real-time ROS~2, addressing both real-time analysis techniques and a wide range of practical enhancements to the ROS~2 software stack proposed over the years. We first provided the necessary background on the internal sclheduling mechanisms of ROS~2 executors and their interactions with various communication middleware, such as the DDS. Later, we classified and discussed existing works on timing analysis for single- and multi-threaded executors.

We then surveyed a broad spectrum of runtime extensions and system-level mechanisms proposed to improve the predictability and performance of ROS~2 applications. These works include custom executor algorithms, runtime mechanisms for heterogeneous computing platforms (e.g., GPUs and accelerators), support for resource-constrained devices via micro-ROS, and profiling and tracing tools that enable systematic experimentation.

We also summarized state-of-the-art techniques for bounding communication delays, including DDS-level modeling and message-filtering strategies. The taxonomies introduced in this survey help to provide a structured view of a rapidly evolving research landscape.

Research directions for future work are manifold. For example, studying mechanisms to enable predictable interaction between ROS 2 and hardware accelerators for AI can yield impactful research. From a theoretical standpoint, it can be interesting to target the impact of model uncertainty (e.g., uncertain execution time estimates~\cite{zini2024search} or uncertain arrival patterns~\cite{vuadineanu2022robust}) on timing metrics. 

However, most importantly, future efforts of the community should enable technology transfer from the research community to the ROS~2 mainline by first achieving a common view of how the new ROS~2 executor should be designed and implemented to guarantee timing constraints.

\bibliography{reference}

\end{document}